\documentclass[11pt]{article}

\usepackage{graphicx}
\usepackage{algorithm}
\usepackage{algpseudocode}
\usepackage{array}
\usepackage{amsmath}

\begin{document}

\title{Articulated Hand Pose Estimation Review}
\author{Emad Barsoum \\ 
        Columbia University \\ 
        \texttt{eb2871@columbia.edu}}
\date{}
\maketitle

\begin{abstract}
With the increase number of companies focusing on commercializing Augmented Reality (AR), Virtual Reality (VR) and wearable devices, the need for a hand based input mechanism is becoming essential in order to make the experience natural, seamless and immersive. Hand pose estimation has progressed drastically in recent years due to the introduction of commodity depth cameras.

\par Hand pose estimation based on vision is still a challenging problem due to its complexity from self-occlusion (between fingers), close similarity between fingers, dexterity of the hands, speed of the pose and the high dimension of the hand kinematic parameters. Articulated hand pose estimation is still an open problem and under intensive research from both academia and industry.

\par The 2 approaches used for hand pose estimation are: discriminative and generative. Generative approach is a model based that tries to fit a hand model to the observed data. Discriminative approach is appearance based, usually implemented with machine learning (ML) and require a large amount of training data. Recent hand pose estimation uses hybrid approach by combining both discriminative and generative methods into a single hand pipeline.

\par In this paper, we focus on reviewing recent progress of hand pose estimation from depth sensor. We will survey discriminative methods, generative methods and hybrid methods. This paper is not a comprehensive review of all hand pose estimation techniques, it is a subset of some of the recent state-of-the-art techniques.\newline

\end{abstract}


\section{Introduction}
Hand pose estimation and gesture recognition provide a natural input mechanism for Human Computer Interaction (HCI) especially in the area of AR and VR. Nevertheless, they are also important scenarios in which the user cannot touch the computing device such as medical doctor during operation, or someone is eating and want to change the playing song or change the reading page (dirty hand scenario), or 10” experience (TV experience) in which you are far away from the screen, or during a presentation in which you want to change the current slide or highlight a section during the presentation, and many more.

\par Some of the non-vision methods used for hand tracking are using gloves with sensors. The main advantage of using gloves is its accuracy and performance compared to vision based state-of-the-art methods, especially in case of heavy occlusion; however, the gap is getting closer. The problems of using gloves are that they are costly, require calibration and not the most natural way for the user to wear yet another device. An advantage of wearing gloves beside accuracy is that they can provide a force feedback.

\par Other vision based hand tracking uses marker in the hand for easy segmentation and part detection, marker usually can be a colored gloves \cite{Wang09} or painted hand \cite{Tompson14, Sharp15} (used offline to capture ground truth) or any other form of marker. The advantage of this approach is that it solves a lot of the challenges of marker-less hand pose estimation, the marker can be pretty cheap, depend on the marker it might not need calibration and the vision algorithm for the marker usually fast and less complicated than the marker-less pose estimation. However, the main issue is that using marker is not natural or as accurate as glove methods. One big advantage of using marker, is that it can be used to create labeled training data \cite{Tompson14, Sharp15} for the marker-less training algorithm, which as we see later is the most time consuming part and labor intensive.

\par There are a lot of literature on hand gesture recognition \cite{Oka02,Krupka14} and hand pose estimation \cite{Oikonomidis10,Oikonomidis11,Fanello14,Qian14,Tang14,Tompson14,Sharp15,Sun15,Oberweger15}; for hand pose estimation, it seems that recently most literature converged on high level architecture of what the hand pose pipeline should look like. And the focus is to improve the different algorithms in each of the pipeline stages. Also, due to Microsoft Kinect and the introduction of a relatively cheap depth sensor, most papers now focus on RGB-D data from depth sensor to estimate hand pose. Nonetheless, there are some papers that tried to estimate 3D hand pose from 2D image with the help of a hand model or depth estimate of the hand(Fanello14). The problem in extracting hand pose from 2D image is that it is many to one mapping, which mean you can have 2 different 3D hand poses projected to the same 2D pose – using temporal might help in this case by providing some context.

\par As for hand gesture recognition, gross hand gestures has a lot of attention both from academia and industries because it is relatively easier than hand pose estimation and more suitable for user interface (UI) interaction (i.e. Hand gesture in Xbox one, Samsung TV…etc). For UI controls, we just need discrete set of gestures such as grab, pick, hand open, hand close…etc, and motion between any 2 endpoints for scrolling. Hand gestures also might be beneficial to speed up hand pose estimation, for example if we know the current gesture we can reduce the search space on some of the algorithms that uses model or search based techniques.

\par There are a third type that can be consider a subset of hand pose estimation or super-set of hand gesture, which is partial hand pose estimation \cite{Erol07, Oka02}. The idea here is that for some applications, there is no need to the full hand pose estimation, knowing the fingers location and tracking the hand is enough to provide pointing, zooming and other dynamic gestures for the application.

\par Gesture recognition and partial hand pose estimation are usually inferred from the observed data directly, there are heuristics, image based search and machine learning techniques for hand gesture recognition. As for hand pose estimation, there are 3 types of pipeline used: appearance based approach similar to hand gesture which is a regression problem \cite{Tang14,Tompson14,Sun15,Oberweger15}, model based fitting which tries to fit hand model to the observed data \cite{Oikonomidis10, Oikonomidis11} and a hybrid approach which uses both techniques in a single pipeline \cite{Qian14, Sharp15}. Inferred from observed data directly is called discriminative approach and fitting a model is called generative approach.

\par The focus of this paper is on hand pose estimation from depth data. I will discuss some of the hand pose estimation from 2D images and some of the hand gesture techniques, but the in-depth discussion and analysis will focus on articulated hand pose estimation from depth data. Pretty much estimating the 20+ degree of freedom (DOF) parameters of the human hand.

\subsection{Related works}
\label{section:relatedwork}
To the best of my knowledge, the latest comprehensive review work on hand pose estimation and hand gesture was published in 2007 \cite{Erol07}, the focus of \cite{Erol07} paper was primarily vision based hand pose estimation.\newline

\par In \cite{Erol07}, they divided pose estimation into two categories:

\begin{enumerate}
\item Appearance based: inferring the hand gesture or pose directly from the observed visual data without the help of a model. This usually implemented using machine learning (ML) which require a large training data or with Inverse Kinematic (IK) or a hybrid approach.
\item Model based: this approach generate multiple hand model usually called hypothesis and it tries to find the model that best fit the observed data. Pretty much it convert the problem into a high dimension optimization problem, finding the model that minimizing a certain cost function.
\end{enumerate}

\par There are a lot of progress and improvements since 2007 in vision based articulated hand pose estimation, especially from depth sensor. This paper will focus primarily on the state of the art works done post 2007.

\par A more recent paper \cite{SupancicRYSR15} in 2015 focused on comparing 13 hand pose estimation algorithms from a single depth frame and evaluated their performance on various publicly available dataset. Furthermore, \cite{SupancicRYSR15} created a new hand training dataset that is more diverse and complex to existing one. \cite{SupancicRYSR15} focus primarily on comparing the quality of each of the 13 algorithms using a common training dataset, this paper focus on reviewing the latest state-of-the-art hand pose estimation algorithms.

\par There are also older reviews \cite{Pavlovic97, Wu01} for hand gesture; however, those reviews focused mainly on gesture recognition and not pose estimation which is a more challenging problem.

\subsection{Outline}
\label{section:outline}
The remaining of this paper will be organized as follow, in section (\ref{section:overview}) we will provide a high level overview of hand pose estimation pipeline and the different hand pose estimation architectures, we will also discuss why it is still a challenging problem. Next, each of the stage shown in section (\ref{section:overview}) will have their own section for in depth analysis and comparison between latest state-of-the-art. Therefore, next section will be segmentation section (\ref{section:segmentation}) which will focus on the various hand segmentation algorithms, followed by the initializer section (\ref{section:initializer}) which focus on appearance based hand pose estimation methods, then followed by the tracking section (\ref{section:tracking}) which focus on the model base hand pose estimation. 

\par Recently, there was 2 deep learning explorations for hand pose estimation that do not fit in the mentioned layout, so following the tracking section is the deep learning section(\ref{section:deeplearning}), which focus on the 2 deep learning papers for hand pose estimation. Next, we will discuss the current state-of-affair of hand pose dataset in section (\ref{section:dataset}), followed by current vision based hand pose limitations in section (\ref{section:challenges}), then followed by future directions in section (\ref{section:future}). Lastly, we conclude our findings in section (\ref{section:conclusion}).

\section{Pose estimation overview}
\label{section:overview}
Vision based pose estimation is the process of inferring the 3D positions of each of the hand joints from a visual input. Although, hand pose estimation is similar in concept to human body pose estimation and some of the hand pose algorithms are inspired or taken from body pose estimation methods, there are subtle differences that make hand pose estimation more challenging. Such as the similarity between fingers, dexterity of the hand and self-occlusion.\newline

\begin{figure}[H]
\centering
\includegraphics[scale=0.45]{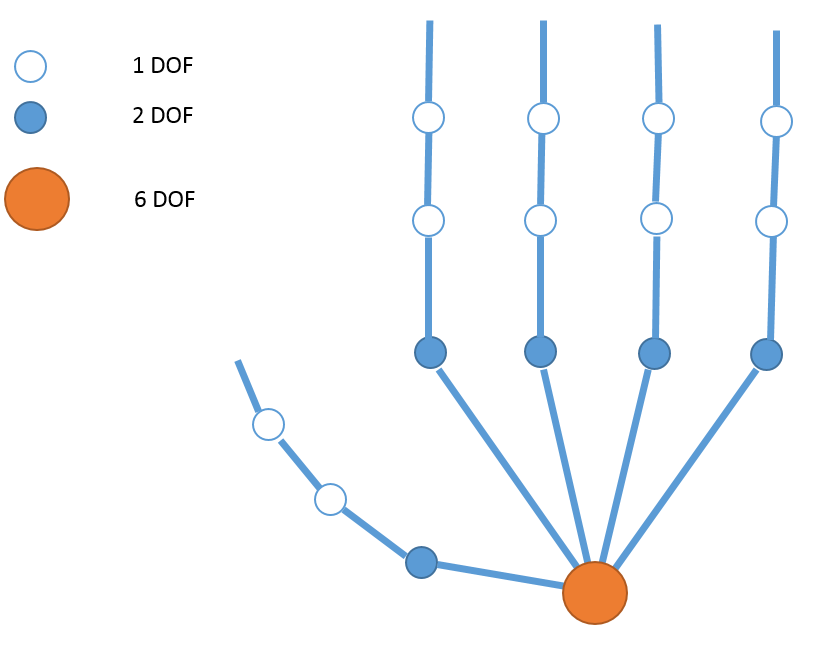}
\caption{An example of hand model with 26 DOF.}
\label{figure:handdof}
\end{figure}

Here some of the challenges for hand pose estimation, the below list assume a non-wearing glove single hand. For more challenging cases are addressed in section (\ref{section:challenges}) which discuss current hand pose estimation limitations.

\begin{enumerate}
\item \textbf{Robust hand segmentation:} while this might seem to be a solved problem, segmenting the hand reliably under unconstrained condition is a difficult task. Hand segmentation is crucial for the quality of the hand pose estimation, most of the reviewed techniques in this paper, their success depend heavily on a good hand segmentation. 
\item \textbf{Degree of freedom (DOF):} hand pose have 20+ DOF that need to be recovered, which is an extremely difficult problem, shown in figure (\ref{figure:handdof}).
\item \textbf{Hand shape:} not all hands are the same, they vary from one person to another. The need to estimate or learn hand shape add more challenges.
\item \textbf{Self-occlusion:} in a lot of hand poses, the occlusion come from fingers occluding each other, which make estimating the pose difficult.
\item \textbf{Speed:} To estimate hand pose, we are dealing with high dimensionality, huge amount of data and complex algorithms. Most algorithms on hand pose estimation are not fast enough for the task that they are trying to solve and some of them require a high end PC or GPU to run in real-time. Furthermore, the amount of time taken by the hand pose estimation algorithm is an added latency to the hand input, in some cases the estimated hand pose wont match current hand pose due to latency.
\end{enumerate}

\subsection{Pose estimation pipeline}

Vision based mark-less hand pose estimation has improved drastically in recent years, the two approaches used for hand pose estimation are discriminative approach and generative approach. Discriminative approach is an appearance based approach, which mean it infer the hand pose from the input data directly. Generative approach uses a hand model and tries to fit the hand model to the observed data.

\par Some hand pose pipelines use discrimininative approach only \cite{Athitsos03, Tang14, Tompson14, Oberweger15, Sun15}, others use generative approach only \cite{Oikonomidis10, Oikonomidis11}, and another use a hybrid approach that combine both discriminative and generative method in a single pipeline \cite{Qian14, Sharp15}. Also, \cite{Tompson14} uses discriminative approach for the hand pose pipeline and generative approach to generate the ground truth training data in order to train their discriminative pipeline. Training data is one of the biggest bottleneck for hand pose estimation, the in-depth discussion of hand tracking dataset is in section (\ref{section:dataset}).

\begin{figure}[H]
\centering
\includegraphics[scale=0.45]{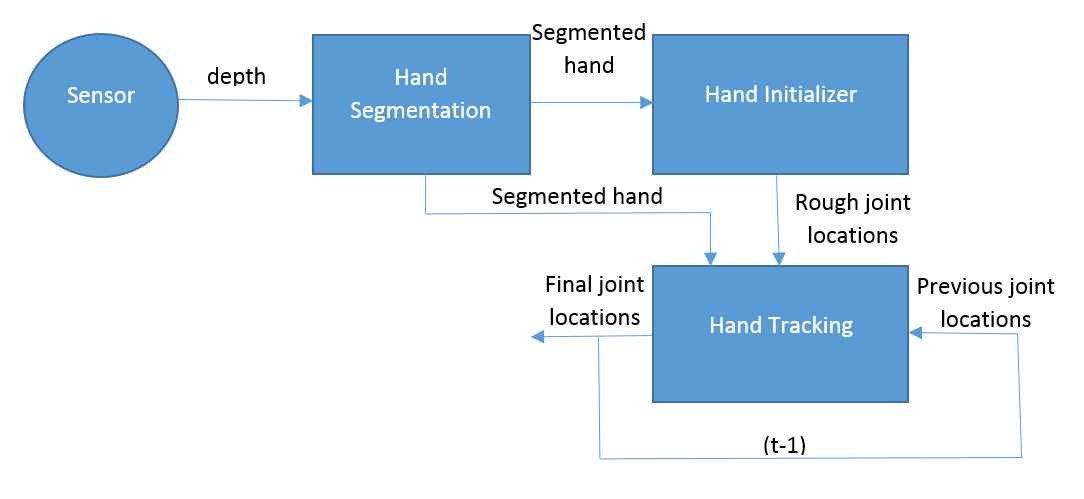}
\caption{A hybrid hand pose estimation pipeline.}
\label{figure:disgenpipeline}
\end{figure}

Figure (\ref{figure:disgenpipeline}) is a high level architecture of a hybrid hand pose estimation pipeline. The initializer stage help to bootstrap and recover the tracking stage in case of tracking failure or during the first frame. The tracking part in this pipeline is the most expensive part in term of compute, it tries to find a hand model that explains the observed data. This pipeline is the most robust but the most costly in term of compute.

\begin{figure}[H]
\centering
\includegraphics[scale=0.45]{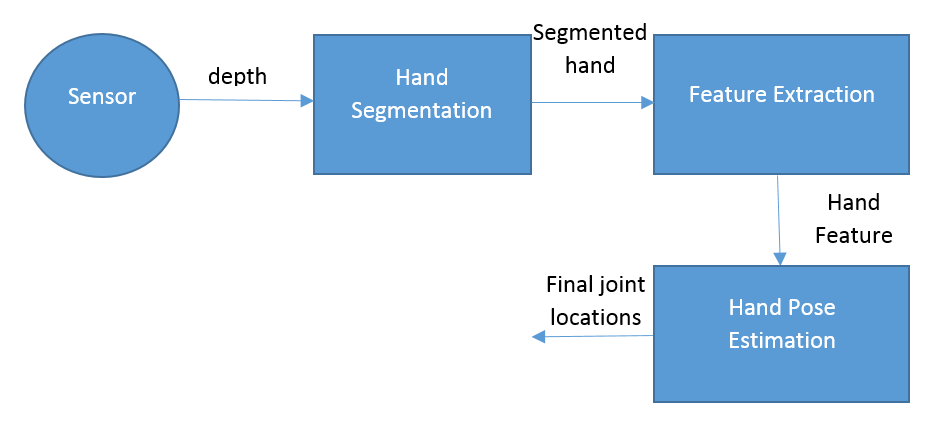}
\caption{Discriminative hand pose estimation pipeline.}
\label{figure:dispipeline}
\end{figure}

Figure (\ref{figure:dispipeline}) is a high level architecture of discriminant hand pose estimation pipeline. This pipeline usually use machine learning (ML) and require a lot of data. One of the main disadvantage of this pipeline is that it does not take into consideration previous result. So the output can be jittery, this can easily be fixed by smoothing the output with the previous output.

\begin{figure}[H]
\centering
\includegraphics[scale=0.45]{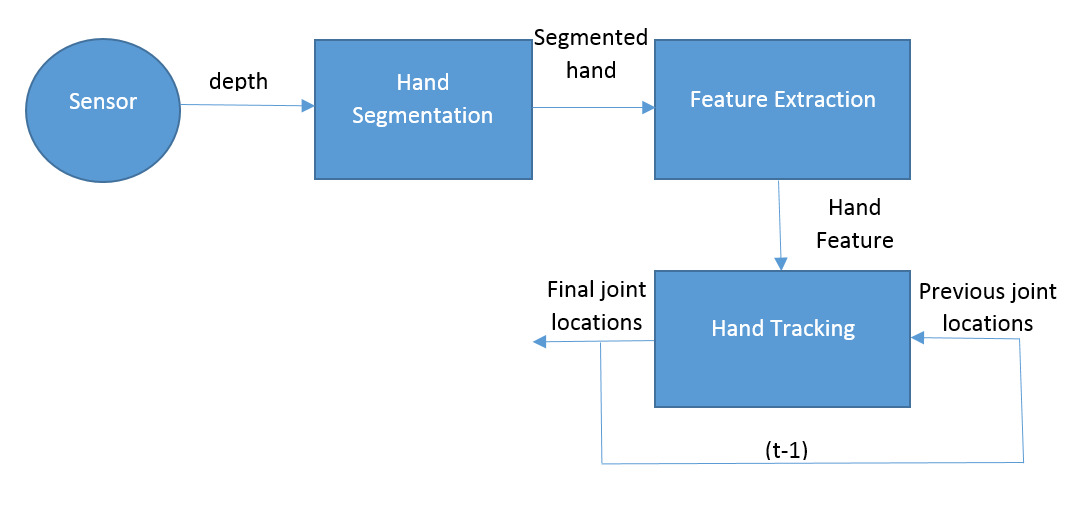}
\caption{Generative hand pose estimation pipeline.}
\label{figure:genpipeline}
\end{figure}

Figure (\ref{figure:genpipeline}) is a high level architecture of a generative hand pose estimation pipeline. This pipeline tries to find a hand model that explain the observed data. The main issue of this pipeline is that it does not recover from tracking failure, it assumes that the hand pose changes between frames are minimum.

\begin{figure}[H]
\centering
\includegraphics[scale=0.45]{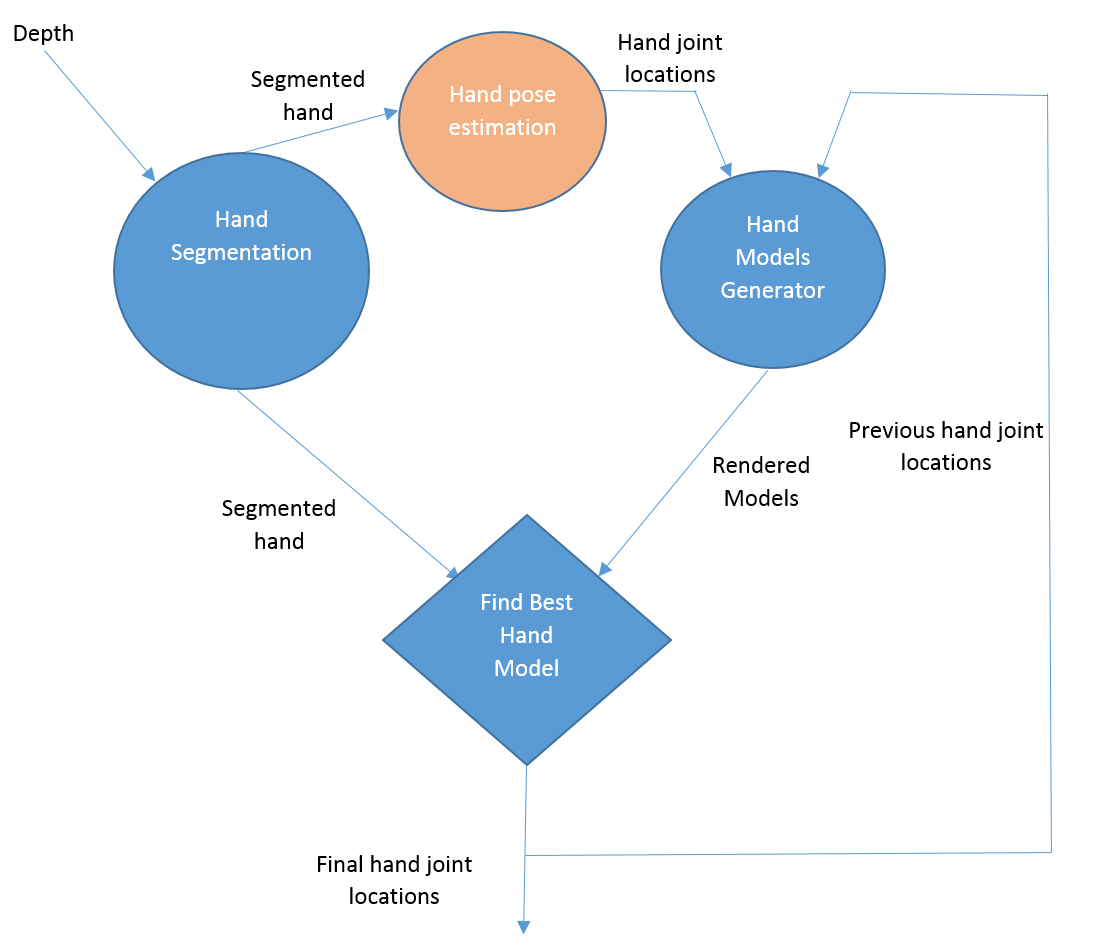}
\caption{Inside the hand tracking stage, the orange circle is only when the initializer is present.}
\label{figure:trackerpipeline}
\end{figure}

Figure (\ref{figure:trackerpipeline}) shows model based hand tracking, the orange piece is only available if we have an initializer stage, otherwise it does not exist. The hand model generators generate multiple hand models around previous frame hand joint result and the initializer output. Those hand models are called hypothesis and they are the input to the optimizer. The optimizer (called "Find Best Hand Model" in the diagram) tries to find which hypothesis explains the observed data the best using a cost function that measure the discrepancy between the observed data and the hand model.

Next, we will go in depth for each of the hand pose pipeline stages. One advantage of having well defined pipeline is that we can pick and choose the algorithm used for each stage without impacting the rest.

\section{Segmentation}
\label{section:segmentation}
One of the disadvantages of using model fitting, is that it is sensitive to segmentation. If the hand segmentation is not accurate the tracking part of most of the techniques mentioned in this paper will fall apart. This is why hand segmentation is crucial aspect for the success of hand pose estimation and it needs more attention.

\par Although hand segmentation might seem an easy problem at first, segmenting the hand in an unconstrained environment is still an unsolvable problem. Here some of the challenges for hand segmentation:

\begin{enumerate}
\item Hand does not have distinct features similar to human faces.
\item Hand is a non-rigid body part, which mean for each pose the shape of the hand is different.
\item Depend on the hand pose, the shadow in the hand can change.
\item For real-time hand tracking pipeline, hand segmentation is the first stage in the pipeline and it needs to be extremely fast in order for the rest of the pipeline, which is more computational intensive, to fit within the real-time constrains.
\item Using Machine learning approach, the most challenging part is to have good coverage for the non-hand cases. Which is extremely difficult.
\end{enumerate}

\begin{figure}[H]
\centering
\includegraphics[scale=0.45]{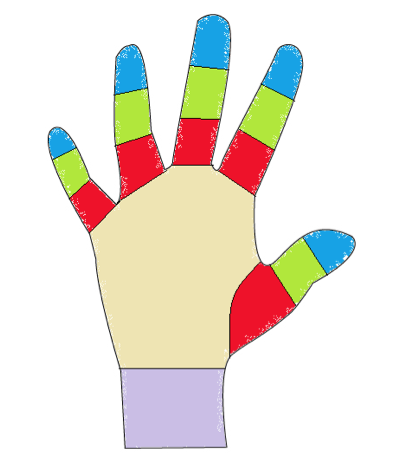}
\caption{Example of hand parts classification.}
\label{figure:handpart}
\end{figure}

Figure (\ref{figure:handpart}) shows hand parts classification output, the output of the ML hand segmentation can be binary classifier (hand or no hand) or multiclass classifier (no hand or hand parts). 

As shown next, most of the current hand segmentation algorithms have some assumptions and added constrains. 

\subsection{Color or IR skin based segmentation}
\par A lot of literature focused on skin based detector for face recognition \cite{Yang02} and hand tracking \cite{Oikonomidis10, Oikonomidis11, Pajdla04}. Skin color detection is attractive for hand segmentation due to its speed, simplicity and the uniqueness of human skin color. Skin detector is usually implemented heuristically, probabilistically or using Machine Learning. 

\par Heuristic methods are based on a color space or combination of color spaces \cite{Yang02, Chai98, Jones99, Saxe96}, the preferred color space is the one that separate chrominance channels from the luminance channel in order to be resilient to illumination changes, \cite{Terrillon00} did comparative study for different skin color models for human face detection in color image which also apply to hand segmentation.

\par For probabilistic methods, the idea is to create a probability distribution that provide the probability for each pixel in the image if it is a skin or not. \cite{Oikonomidis10, Oikonomidis11, Pajdla04} implemented a Bayesian classifier, they used YUV422 color space and ignored the Y channel, which corresponding to illumination, in order to reduce illumination dependency and the amount of data. \cite{Pajdla04} involved training phase and an adaptive detection phase, the training phase trained offline on training dataset and the adaptive phase combine the prior probability from the training phase and the prior probability from the previous $N$ frames to cope with illumination changes.

\par For the machine learning methods, the idea is to train a machine learning algorithm on the input image to distinguish between skin and non-skin area. \cite{Fanello14} trained a random decision forest on Infrared (IR) signal to infer depth from IR skin tone. The training data was capture with the help of a depth sensor registered to the IR sensor, with the assumption that the hand is the closest object to the camera. With the tagged data they run a random decision forest in order to infer depth value from IR skin tone. The tagged data was for each IR frame there is a corresponding depth frame in which the skin pixel has depth and the non-skin pixel has zero depth. There are 2 advantages for this approach:

\begin{enumerate}
\item IR signal is the same under most lighting conditions, which mean the variation of illumination problem in color image does not apply here.
\item They infer depth for each skin pixel not only skin or not skin, which provide more data for hand tracker.
\end{enumerate}

\par While skin based hand segmentation is attractive, it suffers from a lot of problems that make it insufficient for general purpose hand pose estimation:

\begin{enumerate}
\item For color image, even using chrominance channels only, is not sufficient to protect against illumination changes.
\item Skin color detector assumes that no other object in the scene have the same color, which is not true. Even for IR skin tone detector, there are some objects with similar IR level as the human skin.
\item If the person is wearing a short sleeve, skin color detector will segment the rest of the arm which might break hand tracking (Hand model used in \cite{Sharp15} include part of the arm to work around this issue).
\end{enumerate}

\subsection{Temperature based segmentation}
In order to provide a robust hand segmentation that work across different lighting conditions and cluttered background, \cite{Oka02} segment the hand from a passive IR image using a thermal vision infrared camera. The idea is that normal body temperature is constant, so using thermal imaging, \cite{Oka02} segmented the hand with a single threshold that matches body temperature.

\par Although, this method work under different lighting condition and busy background, it assumes that body temperature is constant. Which might not be the case, if someone is sick, or his or her body temperature is a little off because of the weather. Also, it assumes that no other object in the scene have the same temperature as the human body.

\subsection{Marker based segmentation}
In order to increase the robustness and speed of the hand segmentation, \cite{Wang09} segmented the hand using a colored gloves. The glove actually provided a unique color for each part of the hand to help not only the segmentation part but the pose estimation part also. While this approach worked in \cite{Wang09} scenario, it assumed that no other object in the scene have the same color as the glove. Also, wearing a glove in order to do hand tracking is not natural for natural user interaction.

\par Instead of wearing a glove, \cite{Tompson14,Sharp15} colored the actual hand in order to segment and estimate part of the hand. Although, we shouldn't expect people to color their hand in order to use their hand as input mechanism and even with colored hands it still have the same issues as the glove method. \cite{Tompson14,Sharp15} used the colored hand to generate training data only, and then they used machine learning algorithm to train on the generated data. Manually tagging the hand is cumbersome, error prone and does not scale, so painting the hand is a good solution to automate the tagging process.

\subsection{Depth based segmentation}
\par Depth image provide the depth value at each pixel in the scene. One of the big advantage of the depth data is that because we know how far the hand is in the scene, we can roughly estimate heuristically a bounding box around the hand regardless of how far or how close the hand is from the camera. In essence, depth data is hugely beneficial in writing a scale invariant detector.

\par For depth based hand segmentation, one of the assumptions commonly made is that the hand is the closest object to the sensor \cite{Qian14, Oberweger15}, which is not always true especially in office environment where part of the desk is visible to the depth sensor. In \cite{Qian14}, they assumed the hand is the closest object to the sensor and used connected component analysis to find all depth pixels belong to the hand, in order to avoid having the wrist as part of the segmentation, they wore a black band around the wrist to create a depth void.

\par \cite{Oikonomidis11} segmented the hand using both skin based approach from a color image and a depth based approach from a depth sensor. They used the depth data to limit the search space of the hand location, and they used skin color detector from \cite{Pajdla04} to segment the actual hand. They limited the search space to be within of +/- 25 cm from the previous frame.

\subsection{ML based segmentation}
\cite{Sharp15, Tompson14} used random forest for pixel wise classification of the hand, their algorithm is based on the human pose estimation work from \cite{Shotton11}. In \cite{Sharp15}, they used 2 steps process to segment the hand.

\begin{enumerate}
\item Using the output of Kinect body tracker to provide a rough estimate of the hand position.
\item Kinect body tracker hand position is not always precise and does not work closer than 0.5 meter from the sensor. So the second step is Machine learning (ML), a pixel wise classifier from \cite{Shotton11} that classify pixels that belong to the hand from those belonging to the forearm or background.
\end{enumerate}

\begin{figure}[H]
\centering
\includegraphics[scale=0.40]{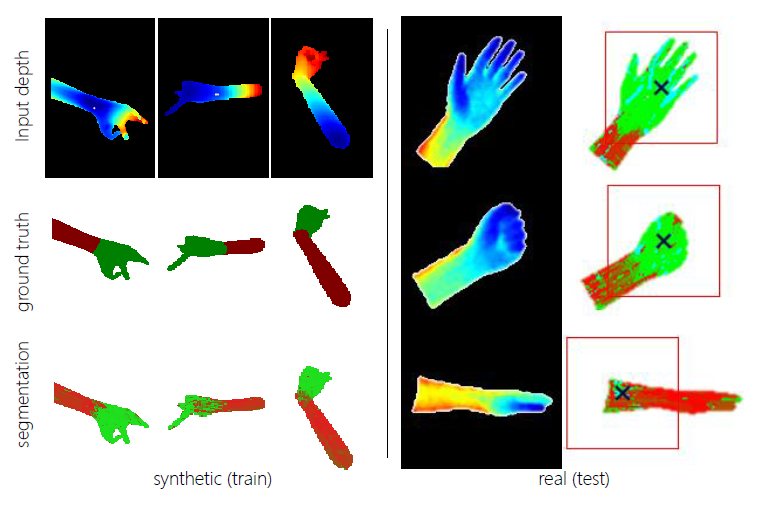}
\caption{From \cite{Sharp15} Hand segmentation:left training data, right test data and last row shows failure case}
\label{figure:seg-sharp}
\end{figure}

\begin{figure}[H]
\centering
\includegraphics[scale=0.40]{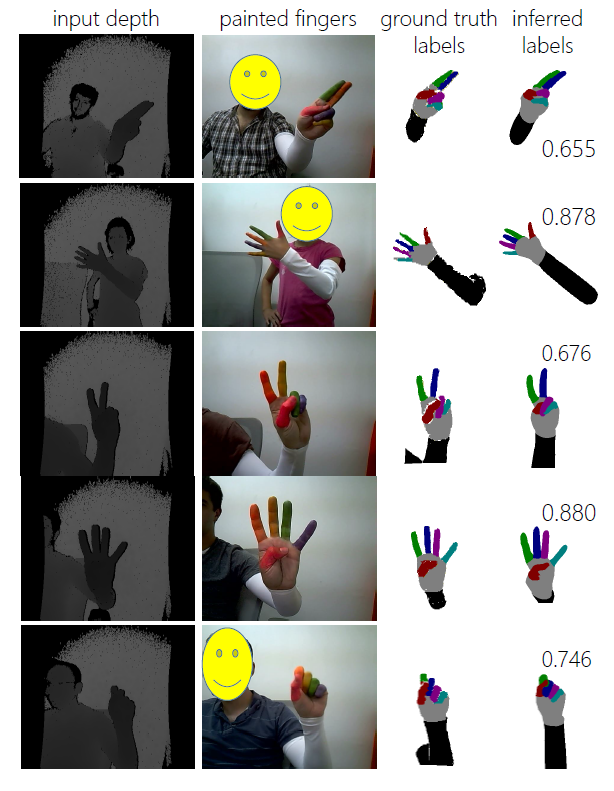}
\caption{From \cite{Sharp15} Hand segmentation ground truth}
\label{figure:seg-sharp-gt}
\end{figure}

In order to automate the process of tagging segmented hand, \cite{Sharp15} capture a video sequence of a painted hands from a Time-of-Flight (ToF) depth sensor, and from a calibrated color camera that is registered to the depth sensor. Then a semi-automatic color segmentation algorithm applied to the captured data in order to produce pixel wise ground truth for finger, palm and forearm. As shown in \ref{figure:seg-sharp-gt}, each finger is painted with different color, and the palm also is painted with a different color. All the participants are wearing long sleeve with uniform white color. 

\section{Initializer}
\label{section:initializer}

The function of the initializer is to infer the most likely hand pose or poses (called hypothesis) that explain current input data. Its main purpose is to help recovering tracking failure and provide estimate of the initial set of hand poses for the tracker in order to constrain the optimizer search space. The better the initializer in estimating the hand pose, the less work and less compute is needed by the tracker to fine tune the final result. Hand pose estimation algorithms in the initializer stage are appearance based techniques (discriminative), which mean they estimate hand pose based on the input frame directly without a hand model to fit.

\par Most discriminative hand pose estimation and gesture recognition algorithms can be used as initializer in the hand pose pipeline described in this paper.

\par Most of the works done in the initializer can be categorized into four different categories:

\begin{enumerate}
\item Heuristics: \cite{Qian14} used heuristics to estimate finger tip locations and palm direction, then used those to estimate the final hand pose.
\item Inverse Kinematic (IK): using hand IK \cite{Tompson14} to estimate hand joint locations.
\item Machine learning: use ML \cite{Keskin12, Sharp15} to estimate hand pose directly from the data.
\item Image Retrieval: \cite{Athitsos03} treat hand pose estimation as image search from a large database of hand poses.
\end{enumerate}

\subsection{Heuristics}
Heuristic techniques are the least reliable method, because they are usually based on a lot of assumptions and specific hand scale.

\par To find finger tips \cite{Qian14} used the extreme points in the geodesic distance for both the 2D XY plane and 1D Z direction. \cite{Qian14} tried to use the 3D point cloud \cite{Baak13, Plagemann10} instead of the 2D XY and 1D Z, however, this approach did not work well with fingers. Each of the top extreme points are considered finger tip proposals, the next step is to find which one is a real fingertip and which one is not.

\par To evaluate each finger tip proposal, \cite{Qian14} grew a finger segment for each finger tip proposal then checked if its geometry is similar to a finger or not. Finger geometry similarities are done using heuristics and template matching. After the evaluation, the direction of each finger tips is estimated using principal component analysis (PCA). Then, all finger segments are removed from the 3D cloud, the remaining blob is the palm, the direction of the palm is also estimated using PCA.

\par To estimate the hand pose, \cite{Qian14} used finger tips, finger directions and palm direction as constrains. From forward kinematic, \cite{Qian14} derived finger tips, finger directions and palm direction from the hand model. Therefore, the optimal hand pose is the one that minimize the delta between the observed finger tips, finger directions and palm directions, and the model finger tips, finger directions and palm directions. Because the finger identity is not known, \cite{Qian14} enumerate all possible finger combinations and select the one that return the minimum cost function.

\subsection{Inverse Kinematic (IK)}
Inverse kinematics (IK), used in hand pose estimation, is simply the solution to a non-linear hand kinematic equations or objective function based on a certain hand model and the end effectors (such as hand joints or finger tips). Hand model in this context is rigid bodies connected via joints. The end effectors are the estimated 3D joint and/or finger tips locations in the depth frame. So in essence, from a set of estimated 3D joint locations, we generate a set of non-linear equations based on the kinematic of the hand model and its constrains, then we try to find a solution. Or a cost function that evaluate how the model align to the observed data. 

\par The non-linear equations do not have close form solution for a complex structure such as hand model, so in order to solve the equations we need to use optimization techniques. The solutions of these equations are the joint configurations for the hand model which is usually the 3D coordinate of each joint. A review to different IK techniques is given in \cite{Unzueta08}.

\par \cite{Tompson14} estimated the joint locations from a heatmap generated by a trained convolution neural network (ConvNet) from a single depth frame, their ConvNet is discussed in detail in section (\ref{section:deeplearning}). Then, they used IK to recover the pose.

\par The heatmap generated from \cite{Tompson14} ConvNet contains 3D or 2D feature points corresponding to the hand joint in the depth image. The $(x,y)$ are the coordinate of the feature point in the depth image and the $z$ is the actual depth value, 2D feature points if the depth value is zero. Using those feature points they minimized an objective function to align the hand model to the inferred features. Equation (\ref{eq:ik}) shows the objective function used in \cite{Tompson14}:

\begin{equation}
\label{eq:ik}
\begin{split}
&f(m) = \sum\limits_{i=1}^n \left[ \nabla_{i}(m) \right] + \Phi(C) \\
&\nabla_{i}(m) = \begin{cases}
                  \Vert (u,v,d)_{i}^{t} - (u,v,d)_{i}^{m} \Vert_{2}, & \text{If $d_{i}^{t} \neq 0$}.\\
                  \Vert (u,v)_{i}^{t} - (u,v)_{i}^{m} \Vert_{2}, & \text{otherwise}.
                \end{cases}
\end{split}
\end{equation}

Where $(u,v,d)_{i}^{t}$ feature position $i$ from the heatmap and $(u,v,d)_{i}^{m}$ is the model feature position $i$ from current pose estimate. And $\Phi(C)$ is a penalty constrain.

\par To find the best model that align with the observed feature points, \cite{Tompson14} used Particle Swarm Optimization (PSO) algorithm. The advantage of PSO is that it can be parallelized and it is resilient to local optima. 

One problem of IK is that it does not perform well with occlusion, such as joints that are not visible to the camera.

\subsection{Machine Learing (ML)}
In this technique, we can turn the initializer problem into a regression problem, train ML on an input data to predict the hand pose. However, regression on high dimension data is difficult in practice \cite{Sun12, Keskin12, Sharp15}. \cite{Keskin12, Sharp15} split the regression problem into two sub-problems called levels or stages, first level predict global features, called Global Expert Network (GEN) in \cite{Keskin12}, and second level predict local features, called Local Expert Network (LEN) in \cite{Keskin12}. This split is also called coarse to fine tune \cite{Tang14, SupancicRYSR15, Sun15}.

\par In \cite{Sharp15}, they used discriminative ferns ensembles \cite{Krupka14} for the first level, and decision jungle \cite{ShottonJungle13} for the second level:

\begin{enumerate}
 \item Level 1: In this level \cite{Sharp15} used discrimination ferns ensembles to infer the global rotation of the hand. The global rotation is quantized to 128 discrete bins.
 \item Level 2: This level is condition on the output of level 1, there is a classifier for each bin. So in this level, there are 128 classifiers for each of the 128 bins. Decision Jungles was used because its small memory footprint allowed it to scale to 128 classifiers. Level 2 predicts:
 \begin{itemize}
  \item Global translation.
  \item Global rotation.
  \item Pose cluster from one of the following clusters: open, flat, half open, closed, pointing, pinching.
 \end{itemize}
\end{enumerate} 

\subsection{Image Retrieval}
In this category, the problem of finding per image hand pose is treated as a content base image retrieval (CBIR), by simply index a large database of image poses with their corresponding hand pose parameters. Then, for an input image extract its features and find the closest hand pose from the hand poses database that matches the input image features. The closest match is the result.

\par \cite{Athitsos03} created a large database of rendered hand poses where each entry contains the hand pose parameters that generated this hand pose view. This part is a preprocessing part that can be done offline. 

\par Now for an input image, \cite{Athitsos03} find the closest hand pose entry in the database, and return the hand pose parameters associated with this entry. The returned hand pose parameters are the hand pose result for the input image.

\par The problem of this method is that it will require a pretty large database in order to accommodate the analog nature of hand pose. Nevertheless, it can be used as first layer in the machine learning approach.

\section{Tracking}
\label{section:tracking}

\begin{figure}[H]
\centering
\includegraphics[scale=0.40]{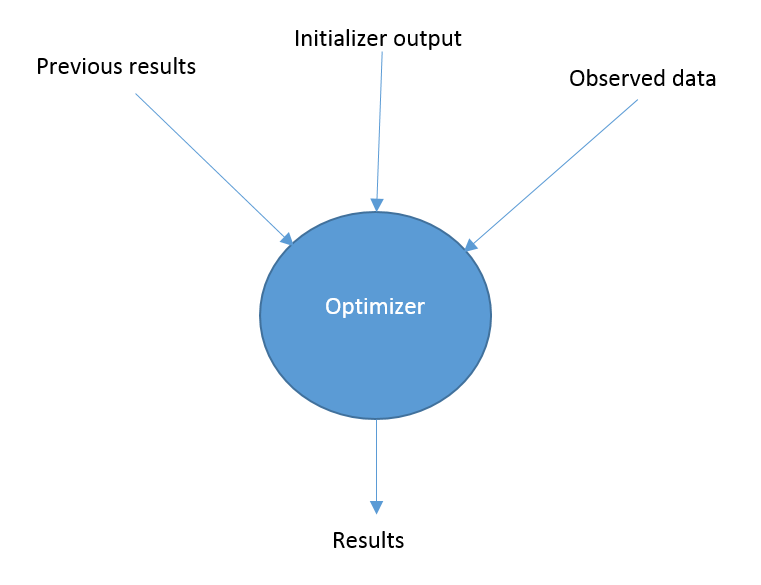}
\caption{Optimizer in model based hand tracker.}
\label{figure:optimizer}
\end{figure}

The goal of the tracking is to estimate the current hand pose from multiple hypothesis and the observed data, these hypothesis are generated from the initializer and previous hand poses, as shown in figure (\ref{figure:optimizer}). The main purpose of the multiple hypothesis is to reduce the number of hand poses that the tracking need to evaluate and constrain the search space. Hand hypothesis is a model of the hand and the evaluation is a cost function that takes a hand model and the observed data as input, and return a single number that measure how close is this hypothesis to the observed data.

\par Due the number of parameters required by the hand model, even with a limited number of initial hypothesis, perturb the model parameters will result a huge number of hypothesis that need to be evaluated. In order, to efficiently search the parameter space of the hand model, most papers use stochastic evolutionary optimization techniques. So for hand tracking to work, we need the following 2 parts:

\begin{enumerate}
\item A good hand model that can express the required hand poses and the corresponding cost function that measure the discrepancy between the observed data and the model.
\item An optimization technique to search the hand model parameter space in order to find the best hypothesis that explain the observed data.
\end{enumerate}

Having a hand model (hypothesis) and a cost function that measure the discrepancy between the hand model and the observed data, the goal of the optimizer is to find the best hypothesis that explain the observed data according to the cost function. 

\par The cost function depends heavily on the selected hand model, the type of observed data and the assumption made to reduce the evaluation of the cost function.

\subsection{Hand model and cost function}
\label{section:model}
Human hand contains many moving parts that interact with each other and provide complex articulation. In order to model the hand, there are a variety of options depend on the balance required between accuracy and performance. The selected hand model and the input signal dictate the design of the cost function.

The characteristic of a good objective function for hand tracking is as follow:
\begin{enumerate}
\item Need to provide a measure of how close a hand model is to the observed data without ambiguity.
\item Need to have constrains against trivial solution that break the kinematic or the anatomy of the hand. Such as overlapping fingers, bone angles that are physically impossible...etc.
\item For real-time system, the objective function is called for each hypothesis. So the evaluation of the objective function need to be fast.
\end{enumerate}

\par Hand model used in literature varies from simple model consistent of basic geometries \cite{Qian14, Oikonomidis10, Oikonomidis11} to a more sophisticated model consistent of full 3D mesh of the hand \cite{Sharp15, Tompson14}. From performance perspective, there are two bottlenecks related to hand models:

\begin{enumerate}
\item To evaluate a hand model with a set of parameters, the hand model need to be rendered first. Which occur for each hypothesis evaluation per frame.
\item Once we have a rendered hand mode, the evaluation itself measure the discrepancy between two 3D point clouds, one from the observed depth and another from the synthetic hand model.  
\end{enumerate}

Both of the above operation are computation expensive.

\subsubsection{Sphere based hand model}
\begin{figure}[H]
\centering
\includegraphics[scale=0.40]{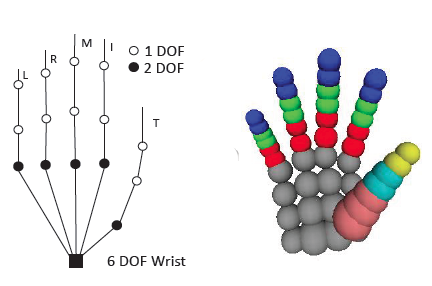}
\caption{From \cite{Qian14} Sphere based hand model and its corresponding DOF.}
\label{figure:spherehand}
\end{figure}

One of the most simple hand model, is sphere based hand model from \cite{Qian14} as shown in Figure \ref{figure:spherehand}. In this presentation, to present a hand, all what we need is the center of each of the spheres and their corresponding radius. \cite{Qian14} adopted 26 degrees of freedom (DOF) similar to \cite{Oikonomidis11, Wang09}.

In order to approximate the hand, \cite{Qian14} used 48 spheres: 6 for each finger except the thumb finger, 8 for the thumb finger and 16 for the palm. The number of spheres chosen for each finger and the palm were entered manually. The sphere size and center were set empirically from the polygon mesh model in \cite{Oikonomidis10}. The model is fixed in size and not adaptable to different hand sizes.

\par One of the huge benefit of the sphere model is that its cost function is relatively fast due to the fact that points on the surface of the hand model are simply points on a sphere which can be evaluated with a single equation.\newline

\par The cost function used in \cite{Qian14} is composed of three terms, shown in equation (\ref{eq:spherefitness}):
\begin{itemize}
\item Align point cloud to model $M$, in order to compute this term in real-time \cite{Qian14} down sampled the point cloud randomly to 256 points (this will affect the number of local optima because of the addition artifacts from down sample).
\item Force the model to lie inside the cloud.
\item Penalize self-collision.
\end{itemize}

\begin{equation} 
\label{eq:spherefitness}
E(P, M) = \lambda \cdot \sum\limits_{p \in sub(P)}D(p, s_{x(p)})^{2} + \sum\limits_{i}B(c_{i}, D)^{2} + \sum\limits_{i}L(s_{i}, s_{j})^{2}
\end{equation} 

Where $P$ is the point cloud and $M$ is the sphere based hand model. $D(.)$ align sub-sampled point cloud to the hand model, for each point cloud $D$ compute the distance between this point to the surface of the closest sphere. $B(.)$ forces the model to lie inside the point cloud, by project each sphere into the depth map, then measure the distance between the actual depth and the projected sphere depth. $L(.)$ penalize self-collision between neighbor fingers, by check if the spheres from both finger overlap or not.

\subsubsection{Geometry based hand model}

\begin{figure}[H]
\centering
\includegraphics[scale=0.40]{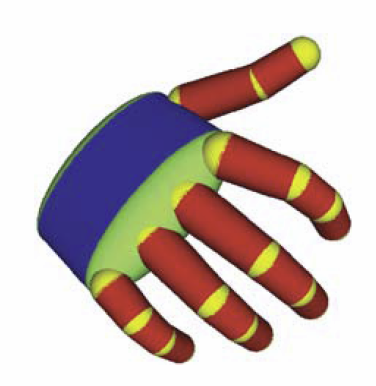}
\caption{From \cite{Oikonomidis10} Basic geometry based hand model.}
\label{figure:geohand}
\end{figure}

\cite{Oikonomidis10} used a hand model based on a number geometry primitives as shown in figure (\ref{figure:geohand}). \cite{Oikonomidis10} uses elliptic cylinder and two ellipsoids for palm, three cones and four spheres for each finger except the thumb, two cones and three spheres for the thumb.

\par \cite{Oikonomidis10} captured multiple images of the hand pose from multiple cameras surrounded the hand, then they projected the 3D hand model to each of the camera view using the camera calibration data. Therefore, the result is $N$ images that capture the real hand pose and another $N$ images that capture the hypothesis. And the goal became how to compare the $N$ observed 2D images with the $N$ rendered 2D images, in order to evaluate the hypothesis.

\par In order to evaluate each hand hypothesis relative to the observed data, \cite{Oikonomidis10} generated descriptors from the observed data and from the synthetic model, then compare them with some objective function define in equation (\ref{eq:fitness1}). The hand pose in \cite{Oikonomidis10} was captured with multiple cameras simultaneous, so each capture is a multi-frame capture, the computed descriptors in each of the observed frame are:

\begin{itemize}
\item Segmentation mask of the hand, generated using skin color to segment the hand from background.
\item A distance transform of the edge map of the image. The edge map was computed using Canny Edge Detector\cite{Canny86}.
\end{itemize}

Using the same notation as \cite{Oikonomidis10}, each image $I$, from the multi-frame capture, will have segmentation mask $o_{s}(I)$ and distance transform map $o_{d}(I)$. In order to compute the equivalent mask and map from the hypothesis, \cite{Oikonomidis10} render each hypothesis to each of the cameras using the camera calibration data $C(I)$. Then from the rendered image \cite{Oikonomidis10} generate the same mask and map as the one from the observed data. The segmentation mask for a hypothesis $h$ corresponding to image $I$ is $r_{s}(h,C(I))$ and the distance transform for the same hypothesis corresponding to image $I$ is $r_{d}(h,C(I))$.

\par The hypothesis evaluation used by \cite{Oikonomidis10} is a distance measure between hand pose hypothesis $h$ and the observed multi-frame data $M$, this distance indicate how closely this hypothesis match the observed data. Here the evaluation function used by \cite{Oikonomidis10}:

\begin{equation} 
\label{eq:fitness1}
E(h, M) = \sum\nolimits_{I \in M}D(I, h, C(I)) + \lambda_{k} \cdot kc(h)
\end{equation} 

At high level, equation (\ref{eq:fitness1}) has two terms:

\begin{enumerate}
\item $\sum\nolimits_{I \in M}D(I, h, C(I))$ this term is responsible for measuring how close the hypothesis is to the observed data.
\item $\lambda_{k} \cdot kc(h)$ this term is a penalty term for kinematically implausible hand configurations.
\end{enumerate}

From equation (\ref{eq:fitness1}), $h$ is the hypothesis, $M$ is the observed multi-frame, $I$ is an image in $M$, $C(I)$ is the camera calibration for the camera that captured image $I$, $\lambda_{k}$ is a normalization factor and the sum is over all images in this multi-frame. $D$ from equation (\ref{eq:fitness1}) is defined as follow:

\begin{equation} 
\label{eq:fitness1details}
D(I, h, c) = \frac{\sum o_{s}(I)\oplus r_{s}(h,c)}{\sum o_{s}(I)+ \sum r_{s}(h,c) + \epsilon} + \lambda\frac{\sum o_{d}(I)\cdot r_{s}(h,c)}{\sum r_{e}(h,c) + \epsilon}
\end{equation} 

Where $o_{s}(I)$, $o_{d}(I)$, $r_{s}(h,c)$, and $r_{e}(h,c)$ are the mask and map for both observed image and rendered hypothesis, the term $\epsilon$ is to avoid dividing by zero, the symbol $\oplus$ is the logical XOR which will return zero if both elements match, and the sum is over the entire mask and map.

\par The above cost function satisfy most of the requirements of a good cost function, except the dis-ambiguity requirement. This is due to the fact that it projects the 3D model into a 2D image which lose information. To mitigate that they surrounded the hand with multiple camera in 360 degree fashion, which make it difficult and awkward to use in 3D interaction scenario.

The same author in \cite{Oikonomidis11} implemented similar algorithm with the same hand model but using depth data from a single Microsoft Kinect sensor instead of multiple RGB sensors. The model was rendered into 3D depth cloud instead of 2D projection.

\subsubsection{Mesh based hand model}

Full mesh model of the hand is currently the most accurate model but also the most expensive model in term of computational resource. \cite{Sharp15} used a full mesh model of the hand that include the wrist for their optimization stage; in contrast, \cite{Tompson14} used the accurate model to generate ground truth of the hand poses for their training algorithm, as discussed in section (\ref{section:deeplearning}).

\par
\begin{figure}[H]
\centering
\includegraphics[scale=0.40]{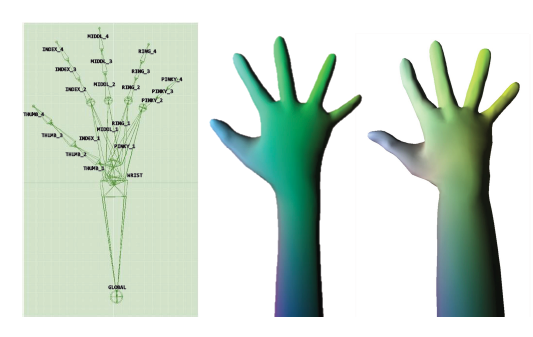}
\caption{From \cite{Sharp15} Mesh based hand model.}
\label{figure:meshhand}
\end{figure}

 Figure (\ref{figure:meshhand}) shows the mesh based hand model used by \cite{Sharp15}. The left image is the kinematic of the hand used in \cite{Sharp15}, center and right are possible hand model generated by standard linear blend skinning from \cite{Taylor14}.

\par For the scoring function, \cite{Sharp15} render each hypothesis (hand model) into a synthetic depth compatible with the real depth data, then compare the synthetic depth with the real depth directly. \cite{Sharp15} define a function that takes the base mesh and the pose parameters $\theta$ of the hand as input, and output the synthetic depth. As shown in equation (\ref{eq:meshscore1}), $r_{ij}$ is the synthetic depth pixel at index $i$ and $j$, for no hand pixel the value equal to the background.

\begin{equation}
\label{eq:meshscore1}
R(\theta; V) = \lbrace r_{ij} \vert 0 < i < H, 0 < j < W \rbrace 
\end{equation} 

And here the scoring function:

\begin{equation}
\label{eq:meshscore2}
E(Z_{d}, R) =  \sum\limits_{ij} v_{ij}\rho(z_{ij} - r_{ij})
\end{equation} 

Where $\rho(.)$ is a truncated linear function kernel.\newline

\par As shown above the cost function is pretty simple, however, the main computation task is the rendering of the hand 3D mesh model for each hypothesis.

\subsection{Optimization}
\label{section:opt}

The tracking part of the hand pose pipeline turns the hand tracking problem into an optimization problem, its goal is to find the parameters of the hand model that minimize the cost function. Because it is an optimization problem, most optimization techniques can be applied. However, the are multiple issues specific to hand tracking optimization that put constrain on the type of optimizer used:

\begin{itemize}
\item The parameter space of the hand model is a high dimension space, usually around 27 dimensions as we will see later.
\item The parameter space contains a lot of local optima.
\item The cost function is expensive in term of compute, because it requires rendering the hand model and compare it to the input 3D point cloud. Both operation are expensive in term of compute.
\end{itemize}

The input to the optimizer is the result of the initializer and previous frame result.

\par In \cite{Erol07} 2007 survey, they reviewed the following optimization techniques used in hand pose estimation: Gauss–Newton method \cite{Rehg94}, Nelder Mead Simplex (NMS) \cite{ouhaddi99}, Genetic Algorithms (GAs) and Simulated Annealing (SA) \cite{Nirei96}, Stochastic Gradient Descent (SGD) \cite{Bray04}, Stochastic Meta Descent (SMD) \cite{Bray04}, and Unscented Kalman Filter (UKF) \cite{Stenger01}.

\par Most recent publications in hand pose estimation use evolutionary algorithm to search the parameters space of the hand model. \cite{Oikonomidis10} in 2010 showed that Particle Swarm Optimization (PSO) can be used efficiently to find the right hand hypothesis that explain the observed data. Current state of the art hand pose estimation uses hybrid approach: PSO to explore the parameter space and be more resilient to local optima, and another algorithm to speed up the convergence. Next we will go in depth to each of the PSO variation used for hand tracking.

\subsubsection{Particle Swarm Optimization (PSO)}

Particle Swarm Optimization (PSO) is evolutionary computation and population-based optimization technique, inspired by social behavior of bird flocking and the field of evolution computation. PSO was introduced in particle swarm optimization \cite{Kennedy95} and described in details in Swarm Intelligence \cite{Kennedy01}. PSO optimize an objective function by keeping track of a population of candidate solutions each called particle, in each iteration (called generation) each of those particles move in the solution space using a simple mathematical formula that depends on the evaluation of the objective function (called fitness) at each particle; the global best solution is updated in each iteration and shared across all particles.

\par In PSO each particle store its current position and current velocity, in addition to the position in which it had the best score of the fitness function. Also, the global best position (solution candidate) across all particles is stored and kept up-to-date. \newline \newline Here the high level steps of PSO:

\begin{enumerate}
\item Initialize the particles at random, the number of particles is given as an input.
\item Evaluate the fitness function or objective function for each particles.
\item Update individual best fitness's and update the global best fitness.
\item Update the velocity and position for each particle
\item Repeat until the global best fitness meet certain threshold or the number of iteration exceed a certain threshold.
\end{enumerate}

Here the equation to update the velocity and position:

\[ v_{i}(t+1) = wv_{i}(t) + c_{1}r_{1}\lbrack \overline{x}_{i}(t) - x_{i}(t)\rbrack + c_{2}r_{2}\lbrack g_{i}(t) - x_{i}(t)\rbrack \]

\[x_{i}(t+1) = x_{i}(t) + v_{i}(t+1)\]

Where the subscript $i$ indicate which particle, $v_{i}(t)$ is the velocity of particle $i$ at iteration $t$, $x_{i}(t)$ is the position of particle $i$ at iteration $t$, $\overline{x}_{i}(t)$ is the best position of particle $i$ at iteration $t$ and $g_{i}(t)$ is the global best solution at time $t$. $w$, $c_{1}$ and $c_{2}$ are parameters provided by the user, and $r_{1}$ and $r_{2}$ are random samples of a uniform distribution between 0 and 1, generated at each iteration.

\par Advantage of PSO are that it is easy to implement, few parameters to tune, easy to parallelize and resilient to local optima. The reason for its resilient to local optima is due to the fact that each particle explore different area in the search space at the same generation.

\par A problem can arise in PSO is particle premature collapse \cite{Kennedy95}, which is caused by premature conversion to a local optima, this usually happen in high dimension data. A mitigation against particle premature, is to have multiple global best fitness one for each sub-swarm particles. So in essence split the particles into a clusters where each cluster has its own global best fitness. Another issue, depend on the cost function PSO might be slow to converge \cite{Qian14}.

\par There are a lot of variation of PSO to mitigate some of its issues and there are a lot of hybrid approaches which is mixing PSO with other techniques.\newline

\par To the best of my knowledge, the first use of PSO in hand pose estimation was introduced by \cite{Oikonomidis10}, \cite{Oikonomidis10} shows that PSO can be used successfully in hand tracking. PSO was used to find the optimal 3D hand model parameters that best fit or explain the observed data. The hand model used has 26 DOF encoded in 27 parameters, so the search space is 27 dimensions and each particle position is 27 dimensions vector encapsulating the parameters of the hand model. Each instance of a hand model is called hypothesis, so the objective of PSO is to find the best hypothesis that explain the observed data. 

\subsubsection{Iterated Closest Point (ICP)}
ICP \cite{Besl92, Rusinkiewicz01} is widely used iterative algorithm to align two point clouds, by fixing one (the observed or scanned point cloud) while keep changing the other until they align. ICP converge fast and suitable for real-time application, but it can easy be trapped in local optima.\newline

Here the high level steps for ICP:

\begin{enumerate}
\item For each point in the source cloud find the closest point in the observed cloud.
\item Estimate the rotation and translation transform between both clouds using mean square error.
\item Apply the transform estimated in the previous step to the source data.
\item Iterate until the mean square error is below certain threshold or it exhausted a maximum number of iteration.
\end{enumerate}

\par \cite{Pellegrini08} generalize ICP algorithm to articulated structure with multiple parts connected with joint or point of articulation. Their experiment showed promising results for both upper body tracking and hand tracking. They represented articulated structure as a tree, where each node is a rigid body part and the edge represent the joint that is connected two body parts. Also, one of the node is arbitrarily selected as the root node. The root node transform is relative to the world coordinate, and all other node transforms are relative to their parents. Each transform between a child node and its parent is constrained by the degree of freedom of this joint. Each body part is presented by a set of points.

\par The extension that \cite{Pellegrini08} provided to generalize ICP for articulated structure, instead of trying to find all closest points from the source to the observed data, they pick a body part first, then find the correspondence of this body part to the observed data while enforce the joint constrains.

\subsubsection{ICP-PSO}
To take advantage of ICP fast convergence, and of PSO resilient to local optima and search space exploration, \cite{Qian14} implemented a hybrid approach that uses both methods to find the closest hand model parameters that matches the observed data, they called the new hybrid algorithm ICP-PSO. The observed that was captured from single Time-of-Flight(ToF) sensor. The result was a real-time hand tracking from a single depth sensor.

\par \cite{Qian14} combined both algorithm by performing ICP for each particle before move to the next PSO generation. Pretty much they used ICP to fast track the convergence at the local particle level.\newline

The hybrid ICP-PSO algorithm used by\cite{Qian14} is shown in Algorithm (\ref{alg:icp-pso}) below:

\begin{algorithm}[H]
\caption{ICP-PSO Algorithm}\label{alg:icp-pso}
\begin{algorithmic}[1]
\Procedure{ICP-PSO}{PreviousHandPose, InitializerResult}
\State $ps\gets GenerateRandomParticles(PreviousHandPose, InitializerResult)$
\For {each generation}
\For {each particle}
\State compute correspondences
\For {m times} \Comment{ICP each particle}
    \State gradient decent on a random parameter
\EndFor
\EndFor
\State k-mean clustering for all particles
\State particle swarm update for each cluster
\EndFor
\State return best particle
\EndProcedure
\end{algorithmic}
\end{algorithm}

As shown above, the k-mean clustering is used to avoid PSO premature convergence.

\begin{figure}[h]
\centering
\includegraphics[scale=0.40]{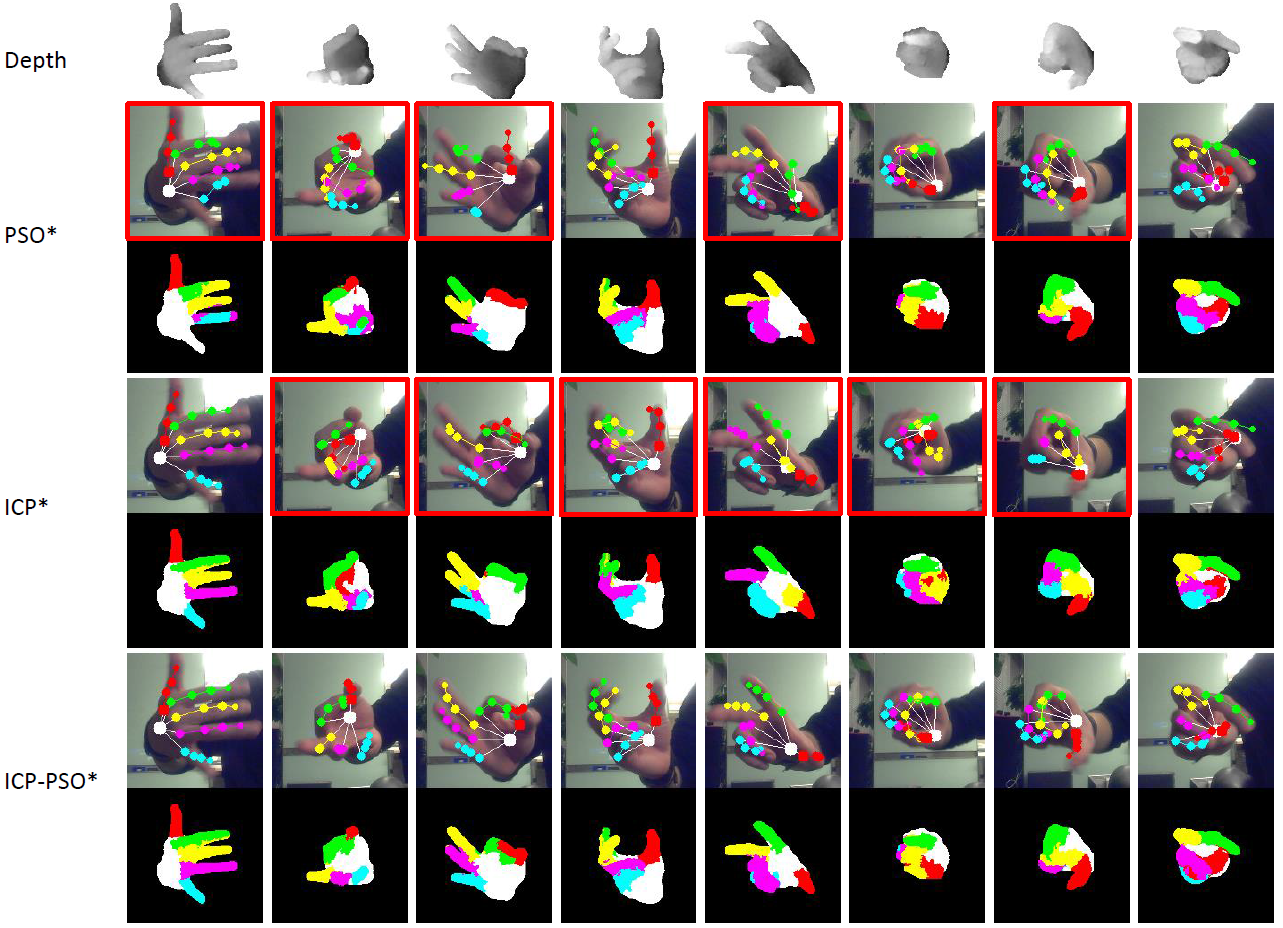}
\caption{From \cite{Qian14} PSO versus ICP versus ICP-PSO}
\label{figure:icp-pso}
\end{figure}

Figure (\ref{figure:icp-pso}), from \cite{Qian14}, shows hand tracking comparison between PSO, ICP and ICP-PSO. The red rectangle highlight the images that has large error. As shown above, ICP-PSO outperform both PSO and ICP alone. To reach a real-time performance \cite{Qian14} used a sparse subset of the input depth data, which will make the data less smooth and will increase local optima from the introduced artifacts. In this setup, ICP-PSO performed better than PSO only or ICP only.	

\subsubsection{Genetic Algorithm (GA)}
\label{section:ga}
Genetic Algorithm \cite{Goldberg89} is a search algorithm inspired by the process of natural selection and genetics in evolution. GA used successfully in optimization problem, especially in problems with little known or large search space. GA depends on a fitness function that evaluate how good each candidate solution in the population is, using techniques similar to evolution such as selection, mutation and crossover, GA will generate a new population from existing one that provide better scoring. After each generation, the search space should move to areas with high value of fitness function, which mean closer to the solution.\newline	

Here the 3 main operations per each generation for GA:
\begin{enumerate}
\item \textbf{Selection:} This step select which individuals will have offspring, the selection is based on a probability associated to each individual. The probability is proportional to individual fitness score.
\item \textbf{Crossover:} Crossover does not occur in every generation, it occurs based on probability. When it occurs, each produced offspring will share parts from 2 parents (2 parts each from different parent will crossover to produce a new individual).
\item \textbf{Mutation:} Mutation also happen with probability, it randomly changes some of the parameters of individual offspring.
\end{enumerate}

The above operations happen at each generation until it converge or exceed a max number of generations.

\par To the best of my knowledge the first use of GA in hand tracking was from \cite{Nirei96}. In \cite{Nirei96} they used 2 optimization techniques:

\begin{enumerate}
 \item GA: To reach close to the solution quickly in the hand model parameters search space.
 \item Simulated Annealing (SA): For local search to find the best model.
\end{enumerate} 

In essence, \cite{Nirei96} used GA to reduce the search space and SA to fine tune the final solution.

\subsubsection{PSO + GA}
\label{section:psoga}
In \cite{Sharp15} they combined features from both PSO and GA, based on a modified version of HGAPSO \cite{Juang04}, for their hand tracking stage. According to \cite{Sharp15} the crucial aspect of PSO is how the next generation of particles are populated from current generation and their scores.

\par Using \cite{Sharp15} terminologies, the algorithm has a population of $P$ particles $\lbrace \Theta_{p} \rbrace_{p=1}^{P}$ and their corresponding score $\lbrace E_{p} \rbrace_{p=1}^{P}$. The main loop is a standard PSO loop, with 2 levels of randomization as follow:

\begin{itemize}
\item Each generation: adjust only fingers, for 50\% of the particles, a random digit is chosen. Its abduction or flexion is adjusted.
\item Each third generation: select 50\% of best performing particles (called elite in \cite{Juang04}) and within these 50\% do the following:
\begin{itemize}
\item Perform local perturbation on 30\% of the particles.
\item Replace 50\% of the particles by drawing new one from a set of poses.
\item Perform GA, splicing or crossover operation, on the remaining 20\% of the particles. By selected a random particles from top 50\% to replace this particle. 
\end{itemize}
\end{itemize}

Most of the parameters tuning, percentage and methods performed above are from intuition and experimentation. So there is a room for more optimal solution. 

\par
\begin{figure}[H]
\centering
\includegraphics[scale=0.50]{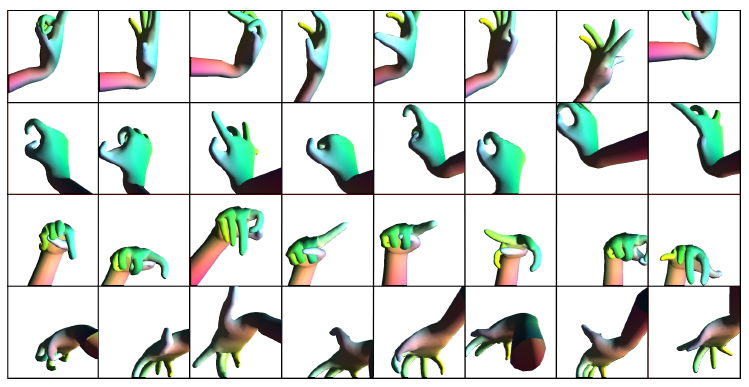}
\caption{From \cite{Sharp15} draw for every third generation}
\label{figure:gapso}
\end{figure}

 In figure (\ref{figure:gapso}) left most column contains the initial pose, and the seven other columns show draw from the every third generation.

\section{Deep learning}
\label{section:deeplearning}

The huge success of deep learning in object recognition and other vision tasks is primarily because the availability of huge amount of images (tagged and non-tagged) and the computational power available today. In case of hand pose estimation, the amount of tagged hand poses from RGB-D available today are still low compared to image recognition dataset. Which might explain why deep learning usage in hand pose estimation is low and did not yet provide the impact that it did for other vision tasks.

\par To the best of my knowledge, there are only two deep learning publications for hand pose estimation \cite{Tompson14, Oberweger15}, both of them are pretty recent 2014 \cite{Tompson14} and 2015 \cite{Oberweger15}. In this section we will discuss both implementations and how they might fit into the general architecture of hand pose estimation.

\par
\begin{figure}[H]
\centering
\includegraphics[scale=0.50]{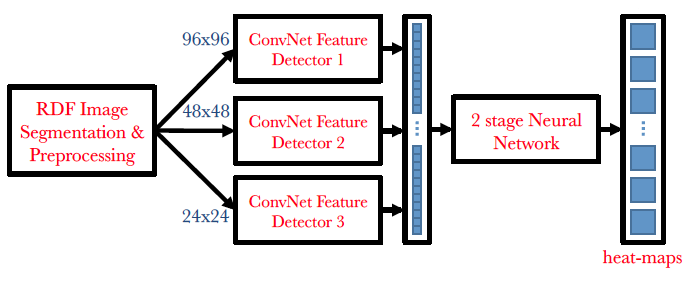}
\caption{From \cite{Tompson14} ConvNet architecture to generate the heatmap}
\label{figure:convnet}
\end{figure}

 In \cite{Tompson14} they trained a convolution neural network (ConvNet) to generate a heat map that highlight all the hand joints visible to the sensor, as shown in figure (\ref{figure:convnet}). The input to the ConvNet is a segmented and preproccessed hand, for the segmentation they used pixel-wise hand classifier similar to \cite{Shotton11}. In the final stage, they used Inverse Kinematic (IK) to find the hand model from the heat map that minimize an objective function.

\par The heatmap is a 2D image that contains the $(x,y)$ coordinate of each visible hand joint and the corresponding depth value is from the depth map, the problem in using IK to find the hand pose based only on the heatmap and not taking into consideration the actual observed depth data, is that it wont be reliable for hidden joints. Nevertheless, the ConvNet stage can be used in the initializer, from section (\ref{section:initializer}), in the general architecture that we discussed previously, and a more robust optimizer can be used to find the final hand pose.

\par An interested work by \cite{Tompson14}, is how they generated their tagged training data. They generated two set of data, one for the segmentation part and another for the ConvNet part.

\begin{itemize}
\item Segmentation training data: Similar to \cite{Sharp15}, they used the painted hand approach to automate the process of generating hand segmentation ground truth.
\item ConvNet training data: they used PSO with partial randomization(PrPSO) \cite{Yasuda10} and high quality hand model (Linear Blend Skinning[LBS]), offline in order to find the best hand pose that explain the depth data based on a modified version of \cite{Oikonomidis11} algorithms. They fine tune the output of the PSO with Nelder-Mead optimization algorithm \cite{Tseng99}. The result of the tracking is the ground truth for the ConvNet, doing the model based tracking part offline means that only quality matter and not performance.
\end{itemize}

\begin{figure}[H]
\centering
\includegraphics[scale=0.50]{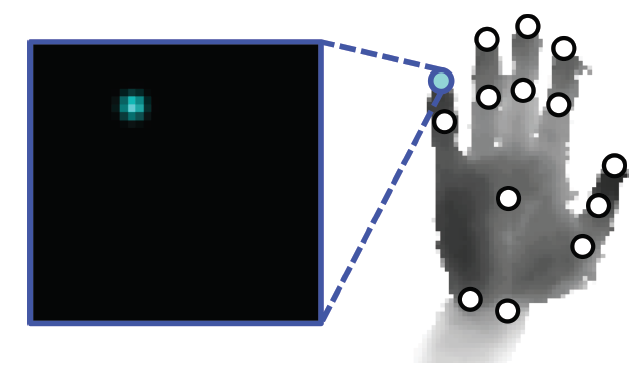}
\caption{From \cite{Tompson14} Heatmap result with feature points overlaid.}
\label{figure:heatmap}
\end{figure}

\begin{figure}[H]
\centering
\includegraphics[scale=0.50]{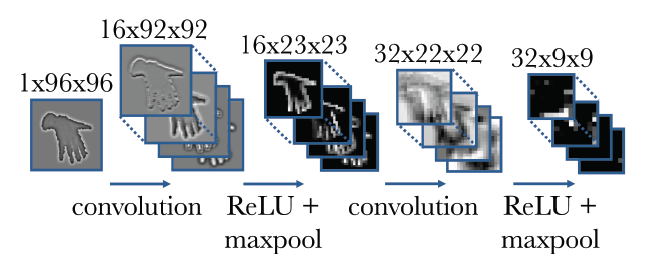}
\caption{From \cite{Tompson14} One of the feature detector ConvNet.}
\label{figure:convnetfeature}
\end{figure}

As shown in figure (\ref{figure:convnet}), the input depth is segmented using a random decision forest (RDF) to generate a depth map that have depth values only on the hand pixels and background value otherwise. Then, the segmented hand is preprocessed, the output of the preproccessing is 3 different fixed resolutions in which each is fed to a ConvNet feature detector shown in details in figure (\ref{figure:convnetfeature}), the output of the detectors is fed into 2 layers full neural network to generate the heatmap. And example of heatmap result is shown in figure (\ref{figure:heatmap}).\newline

\par In \cite{Oberweger15}, they used different approach than \cite{Tompson14} in their Convolution Neural Network(CNN) architecture. They used two networks: one to infer the world position of each hand joint, and the other to fine tune the result, so it is similar to coarse to fine layers from \cite{Tang14, Sun15}. For the segmentation part, \cite{Oberweger15} simply assumed that the hand is the closest object to the sensor, they created a 3D bounding box around the hand, resize it to $128x128$ pixels, then normalized the depth to be in $\left[ -1, 1\right]$ range.

\par One of the contribution by \cite{Oberweger15} in CNN for hand tracking is the incorporation of hand kinematic constrains in the network. Due to the strong correlation between different 3D hand joint locations because of hand kinematic, it is possible to represent the parameters of the hand in a lower dimensional space \cite{WuLin01}. In order to embed and enforce such constrain in CNN, \cite{Oberweger15} added a bottleneck layer with less neurons than needed by full pose representation, this bottleneck layer force the network to learn a low dimension presentation of the hand pose.

\begin{figure}[H]
\centering
\includegraphics[scale=0.50]{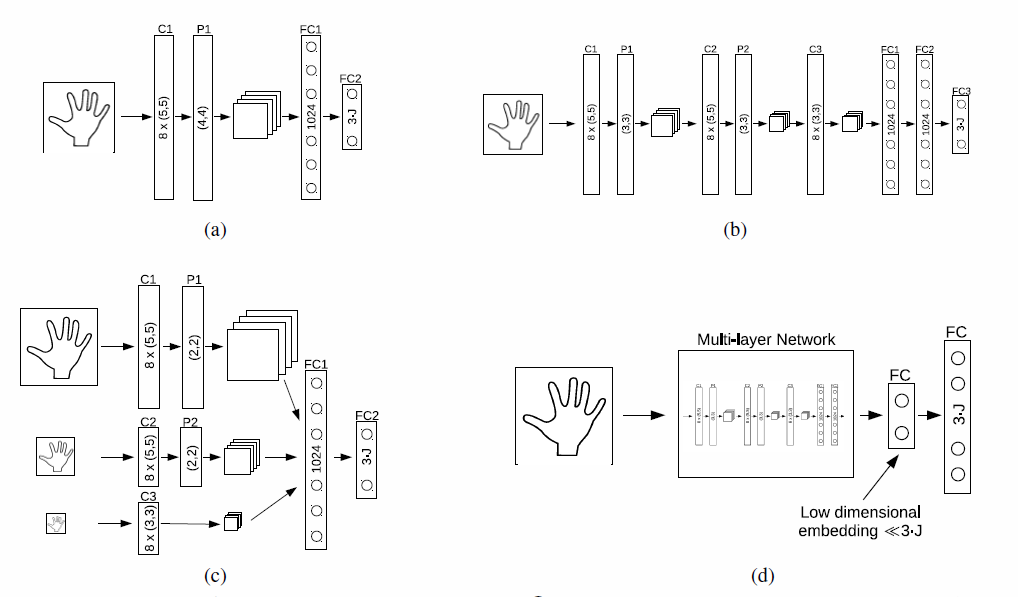}
\caption{From \cite{Oberweger15} different CNN architectures used in stage 1}
\label{figure:cnnbottleneck}
\end{figure}

\par Figure (\ref{figure:cnnbottleneck}) shows the different network architectures tried by \cite{Oberweger15} with the addition of the bottleneck layer. (a) shows a shallow CNN network, (b) shows a deep CNN network and (c) shows the multi-scale network. (d) is the final architecture with the bottleneck layer, the Multi-layer Network is simply one of the networks in (a), (b), and (c). According to \cite{Oberweger15} (c) performed best followed by (b) then (a).\newline

\par Next stage is another network that refine the result from first stage, the architecture of this network is called Refinement with Overlapping Regions(ORRef) by \cite{Oberweger15}. The input to the network is several patches with different sizes centered around joint location from the the first stage. The pooling is done only on the large patches, and the size of the pooling regions depend on the patch size. The reason for no pooling in small patches is accuracy.

\begin{figure}[H]
\centering
\includegraphics[scale=0.45]{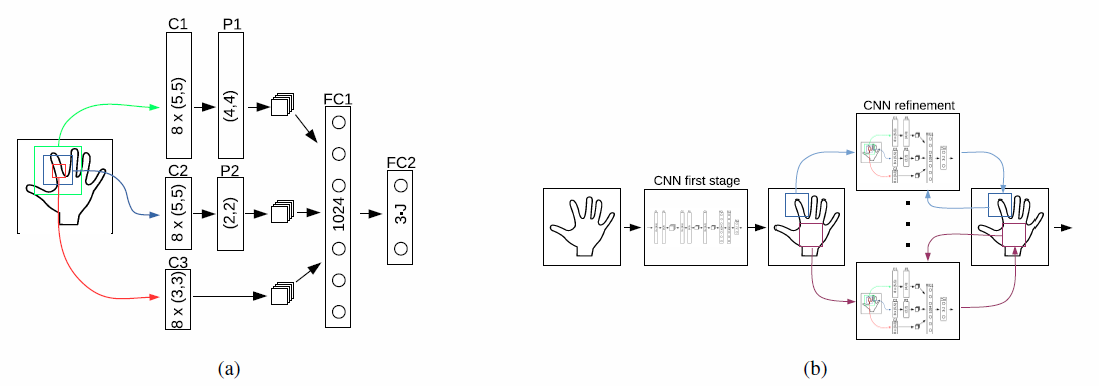}
\caption{From \cite{Oberweger15} CNN architectures used in stage 2}
\label{figure:cnnbottleneck2}
\end{figure}

As shown in figure (\ref{figure:cnnbottleneck2}), there are multiple refinement networks one per joint.\newline

\par Both deep learning approaches discussed in this section are per frame regression of the hand pose joints, they do not take previous frame result into consideration.

\section{Dataset}
\label{section:dataset}
\par Although there are a great progress in recent years for 3D hand pose estimation from single depth camera, the amount of training hand data available publicly are still pretty low. And within those available the dataset are limited in number of subjects and poses as we will see later. One of the main problem in generating tagged hand pose data is that it is time consuming, error prone and difficult to scale the process unless we introduce some form of automation or semi-automation. 
\par Some of the works that we discuss here come up with innovative way to automate the process of tagging hands in images, as follow:

\begin{itemize}
\item For hand segmentation the main method used is coloring the hand as shown in \cite{Sharp15, Tompson14}.
\item For hand pose, \cite{Tompson14, Sun15} used a slow but high quality hand pose estimation pipeline in order to generate ground truth offline, then manually corrected the result.
\item Another approach for hand pose is to use synthetic pipeline to generate a ground truth for hand pose estimation, assuming that the synthetic pipeline is accurate enough to mimic human hand as shown in \cite{RogezSKMR14}. \cite{libhand} is a well known open source library that generate floating synthetic hand pose. Even if the hand synthetic dataset is accurate, we still need a real hand data for verfication and testing.
\end{itemize}

Table (\ref{table:dataset}) shows some of the available non-synthetic hand pose dataset:
\begin{table}[H]
\begin{center}
\begin{tabular}{ |c|p{5cm}|p{5cm}| } 
 \hline
 \textbf{Dataset} & \textbf{Description} & \textbf{Comment} \\
 \hline
 NYU \cite{Tompson14} & 
 NYU dataset contains 72757 training hand pose frames of RGB-D data with ground truth information and 8252 test set. The data are from 3 Kinects camera: a front view and 2 side views. The dataset also contains 6736 tagged depth frames for segmentation. The training data are from one subject and the test data are from 2 subjects.&
 While NYU contains a good set of complex hand pose ground truth, the problem is that the training data, for both hand pose and segmentation, are from one subject only and the test data are from 2 subjects only. Which mean the data is biased. Furthermore, 73K frames is very little.\\ 
 \hline
 ICL \cite{Tang14} & 
 ICL data was captured from Intel’s Creative Interactive Gesture Time of Flight(ToF) Camera. They used \cite{Melax13} to generate the ground truth for each frame than manually refine it. The data is from 10 different subjects with different hand size. Total number of captured ground truth is 20K images, with rotation applied to this data, the final frame count is 180K. & 
 While it is better than NYU data in term of the number of subjects and variety of hand size. It has 2 problems; first, the total number of original data is 22K only, and second, the captured hand poses are not as complex as NYU dataset.\\
 \hline
  MSRA \cite{Sun15} &
  Captured 76500 depth images from 9 subjects using Intel’s Creative Interactive Camera. The ground truth was generated using the optimization method in \cite{Qian14}, then manually corrected. Each subject was asked to do 17 gestures chosen from American Sign Language under large view point and fast motion. &
Although this dataset contains a good number of subjects and gestures, it is not large enough. \\
 \hline
\end{tabular}
\end{center}
\caption{Some of the publicly available non-synthetic hand pose dataset.}
\label{table:dataset}
\end{table}

\section{Limitations and challenges}
\label{section:challenges}
Even with the huge progress in recent years for vision based mark-less hand pose estimation, current state of the art still far away from human level recognition and still does not match the non-vision gloves based approach. Looking at the state of the art hand pose estimation reviewed in this paper, we can see that most of them assume single hand, the background not clutter or busy, no glove on the hand, the hand scale is predetermine and the hand is empty (not holding any object). Some of those constrains might be acceptable in certain cases, however, for general and reliable hand pose estimation, and other cases, we need to address those constrains.\newline

\par Here the list of current hand pose estimation challenges:
\begin{itemize}
\item \textbf{Hand segmentation:} Hand segmentation is still a challenging problem especially with clutter background and sleeve. Short sleeve make it difficult to segment the hand from the wrist up without taking part of the arm. The most promising solution for hand segmentation is machine learning approach such as the pixel wise classification from \cite{Sharp15, Tompson14}, which classify if a pixel belong to a hand or non-hand. The limitation of this approach is that it require a lot of training data.  
\item \textbf{Two hands:} Most of the focus for hand pose estimation is using one hand, and for 2 hands they simply run 2 separate hand pose pipelines. The problem in this case, beside doubling the computation amount, is that it does not work when both hands interact with each other and occlude each other. To the best of my knowledge, non of the publicly available training dataset have 2 hands in a single scene interacting with each other. 2 hands complicate all hand pose estimation pipeline stages.
\item \textbf{Object grabbing:} Grabbing an object by hand is another challenge, there are 2 problems from hand pose estimation perspective when the hand is holding an object:
  \begin{enumerate}
  \item The object will occlude part of the hand, which will make pose estimation more difficult.
  \item If the object is not properly segmented from the hand, it can cause false hand pose estimation because the hand pipeline can treat part of the object as hand part.
  \end{enumerate}
\item \textbf{Scale:} Not all hands are created equal, the shape of the hand vary from one individual to another, especially between kids, women and men. Each have different scale and slight different shape. Most of the publications reviewed here assumed a fixed scale. We need a way to estimate the hand scale and shape or learn them, this can happen at the initializer stage or the hand tracking stage. In hand tracking stage, scale and shape can be added as parameters to the hand model, in essence increase the degree of freedom of the hand model.
\item \textbf{Gloves and sleeves:} None of the literature reviewed in this paper support wearing gloves except when it is used as a marker. All publicly available hand pose dataset are gloves free. Also, sleeve is another problem depend on its materials and how long or short the sleeve is, it will usually affect the segmentation process which in turn will affect all downstream stages.
\item \textbf{Dataset:} Compared to object recognition and classification the number of hand pose dataset available publicly is still pretty low, and most of what is available does not cover some of the challenges mentioned in this list. The lack of wide variety of hand pose dataset make it difficult to compare various hand pose estimation algorithms, \cite{Sun15} created their own dataset because existing one was not complex enough. 
\item \textbf{Degree of freedom (DOF):} Hand kinematic has a high number DOF, which complicate the optimizer and can easily fall to local optima.
\item \textbf{Computation expensive:} Latest state of the art hand pose estimation algorithms are pretty expensive in term of resource and run close to real-time using high end machine.
\end{itemize}

\section{Future directions}
\label{section:future}
In this section, we will focus on what direction merit more focus in order to progress hand pose estimation forward. One of the most important aspect that begin to get attention recently \cite{SupancicRYSR15, Sun15} is the availability of a large publicly tagged hand pose dataset that cover variety of hand poses from different subjects with different hand sizes. The poses need to contain challenging hand poses in a cluttered background.

\par The other important aspect is the direction of the hand algorithm itself. The most promising direction of hand pipeline is to have hybrid approach discriminative and generative in a single pipeline \cite{Qian14, Sharp15}. However, the current bottleneck in this approach in the optimization part which require to search a high dimension space plagued with local optima, and in each evaluation it needs to render a complex hand model. So, the focus should be to improve the initializer in order to make the per frame prediction as good to the final result as possible, that will make the tracking part much simpler.

\par Another part that need attention is hand segmentation, current hand segmentation works in most cases except with complex scene such as cluttered backgrounds. The most promising algorithm for hand segmentation is machine learning based algorithm that classify each pixel.

\par Next, we will discuss each of the above items in depth.

\subsection{Dataset}
There is a huge need for a large training dataset for hand pose estimation and hand segmentation on variety of subjects that can be used for benchmark between algorithms in order to move vision based hand pose quality forward. Furthermore, the complexity of the hand poses in the training set need to cover wide range of complexity and some of the issues that we listed in section (\ref{section:challenges}) such as 2 hands manipulation, wearing gloves, different hand sizes and variety of sleeves.

\par Generating ground truth from synthetic data is a great way to scale up the training and test hand data creation, in addition to bootstrap hand pose algorithms. Nevertheless, synthetic hand data does not replace real data. And it is very difficult to generate synthetic training data for segmentation, due to the need of wide variety of different backgrounds.\newline

\par Here some of the possible improvements for synthetic hand data:

\begin{itemize}
\item The ability to simulate gloves and sleeves with a wide variety of materials.
\item The ability to render wide variety of hand poses with speed and dexterity similar to human hand.
\item The ability to add specific camera noise model and depth artifacts such as for ToF sensor multipath, mixed pixels...etc.
\item The ability to simulate cluttered background.
\item The ability to render hand model in infrared (IR) and color frame. Newer depth sensor support both streams.
\item The ability to simulate 2 hands and their interaction to each other.
\end{itemize}

\par For generating training hand data based on real capture, the 2 promising directions to reduce the time required to tag each frame are:

\begin{enumerate}
\item \textbf{Segmentation training data:} \cite{Sharp15, Tompson14} registered a color camera to the depth camera and painted their subject hands with a specific color. Then, they used a color segmentation algorithms to segment the hand and produce the needed ground truth for hand segmentation. This approach help automate the process of generating training data for hand semgentation which is a tedious and slow task if done manually.
\item \textbf{Hand pose training data:} \cite{Tompson14, Sun15} used a high quality hand model to generate the initial ground truth by running an optimizer on the input data, then manually adjust the result. The advantage of this process is that it is semi-automatic and the only manual work needed is for quality assurance. The use of high quality hand model is not an issue here performance wise, because this process run offline.
\end{enumerate}

\par Another direction that will help moving vision based hand pose estimation field forward is to have competitions similar to what available in object classification, such as PASCAL Visual Object Classes (VOC) \cite{Everingham10} and ImageNet \cite{Russakovsky15}. These competitions in object classification helped spur contest between different universities that caused improvement in image recognition algorithms each year and helped increase the training data for object recognition (now in millions).

\subsection{Hand pose}
If we have a perfect hand initializer, then we wont need a hand tracker, and if we have a perfect hand tracker, then we wont need an initializer. However, in reality we need the initializer for hand tracking loss and the hand tracking to fine tune the initializer estimate. Therefore, the most promising solution for hand pose estimation is the one that uses both discriminative and generative approaches in a single pipeline.

\par Most of the hand pose estimation algorithms reviewed in this paper, still use heuristics and magic numbers from intuition. So there is a room for improvement by finding the optimium values and replace heuristics with more theoritically sound algorithms or machine learning. For example, in \cite{Sharp15} all the percentages selected for the PSO+GA algorithm were from intuition and try-and-error. Therefore, focusing on fine tune to parameters might result better hand pose estimation.

\par Furthermore, the initializer stage need more attention. The focus should be to make the per frame hand pose estimation as close as possible to the final result, that will reduce the dependency on the hand tracking part which is the current performance bottleneck. The reason for that is that the optimizer inside the hand tracking search a high dimension space for a hand model that explain the observed data, this hyperdimension space has a lot of local optima. And in order to evaluate the discrepancy between the hand model and the observed data, the algorithm need to render the hand model then apply a complex cost function to do the comparison. Each of these steps are costly in term of performance and error prone.

\par The 2 promising direction in the initializer stage are: 
\begin{itemize}
\item \textbf{Machine learning (ML):} Multi-layers ML algorithms look promising, by multi-layers we mean coarse to fine tune hand pose estimation \cite{Sharp15, Sun15, Tang14, Oberweger15}. The first layer usually compute global parameters of the hand such as rotation, orientation and location. Given those parameters the second layer infer local hand parameters. 
\item \textbf{Image search:} Content based image retrievel (CBIR) could estimate discrete poses of the hand \cite{Athitsos03, Krupka14}, then we feed this pose to ML algorithm that predict the final hand pose. In concept, CBIR will act as the first layer in multi-layers ML system for hand initializer.
\end{itemize}
 
\par Another important aspect is benchmark, in order to improve on current algorithms, we need to be able to compare various hand pose estimation algorithms together using a publicly available benchmark. Moreover, the benchmark should not be only on the final result, it should also cover each stage.

\par For segmentation and initializer, the most promising approach is machine learning (ML) approach that classify each pixel as a hand or not a hand. Which depend heavily on the availability of large training dataset.

\section{Conclusion}
\label{section:conclusion}
Articulated hand pose estimation based on vision without marker is a challenging and open problem. Hand pose provides a natural interaction in a lot of important scenarios such as TV, Car, 3D manipulation, Virtual Reality (VR) and Augmented Reality (AR). There is a need to have a reliable and robust vision based hand pose estimation in real-time under unconstrained condition, especially with the proliferation of wearable devices.

\par In this paper, we reviewed various hand pose pipelines and reviewed in-depth current state-of-the-art hand pose estimation algorithms, we also looked at each of the hand pipeline stage in detail. Current state-of-the-art almost solved hand pose estimation challenge for a single isolated non-wearing gloves hand with somewhat challenging background. Nevertheless, we are still far away from human level recognition and the ability to infer hand pose in unconstrained environment, especially the 2 hands interaction case and hold an object case.

\par We also show the importance and difficulty to have a large training dataset in order to progress vision based hand pose estimation quality further. Furthermore, we discussed which stage in the hand pose pipeline need more attention and which technique show promising.

\par Articulated hand pose estimation has the potential to revolutionize the way we interact with technology, by making the interaction natural and seamless. And with the recent progress in hand pose estimation, we are getting closer to achieve this goal.

\section*{Acknowledgment}
I would like to thank Professor John Ronald Kender for giving me the opportunity to investigate in-depth hand pose estimation algorithms and gain breadth of knowledge for most vision based hand pose estimation.


\medskip

\begin{thebibliography}{10}

\bibitem{Wang09}
Robert~Y. Wang and Jovan Popovi\'{c}.
\newblock Real-time hand-tracking with a color glove.
\newblock In {\em ACM SIGGRAPH 2009 Papers}, SIGGRAPH '09, pages 63:1--63:8,
  New York, NY, USA, 2009. ACM.

\bibitem{Tompson14}
Jonathan Tompson, Murphy Stein, Yann Lecun, and Ken Perlin.
\newblock Real-time continuous pose recovery of human hands using convolutional
  networks.
\newblock {\em ACM Trans. Graph.}, 33(5):169:1--169:10, September 2014.

\bibitem{Sharp15}
Toby Sharp, Cem Keskin, Duncan Robertson, Jonathan Taylor, Jamie Shotton, David
  Kim, Christoph Rhemann, Ido Leichter, Alon Vinnikov, Yichen Wei, Daniel
  Freedman, Pushmeet Kohli, Eyal Krupka, Andrew Fitzgibbon, and Shahram Izadi.
\newblock Accurate, robust, and flexible real-time hand tracking.
\newblock CHI, April 2015.

\bibitem{Oka02}
Kenji Oka, Yoichi Sato, and Hideki Koike.
\newblock Real-time tracking of multiple fingertips and gesture recognition for
  augmented desk interface systems.
\newblock In {\em Proceedings of the Fifth IEEE International Conference on
  Automatic Face and Gesture Recognition}, FGR '02, pages 429--, Washington,
  DC, USA, 2002. IEEE Computer Society.

\bibitem{Krupka14}
Eyal Krupka, Alon Vinnikov, Ben Klein, Aharon Bar{-}Hillel, Daniel Freedman,
  and Simon Stachniak.
\newblock Discriminative ferns ensemble for hand pose recognition.
\newblock In {\em 2014 {IEEE} Conference on Computer Vision and Pattern
  Recognition, {CVPR} 2014, Columbus, OH, USA, June 23-28, 2014}, pages
  3670--3677, 2014.

\bibitem{Oikonomidis10}
Iasonas Oikonomidis, Nikolaos Kyriazis, and Antonis~A. Argyros.
\newblock Markerless and efficient 26-dof hand pose recovery.
\newblock In {\em Proceedings of the 10th Asian Conference on Computer Vision -
  Volume Part III}, ACCV'10, pages 744--757, Berlin, Heidelberg, 2011.
  Springer-Verlag.

\bibitem{Oikonomidis11}
Iason Oikonomidis, Nikolaos Kyriazis, and Antonis~A. Argyros.
\newblock Efficient model-based 3d tracking of hand articulations using kinect.
\newblock In {\em British Machine Vision Conference, {BMVC} 2011, Dundee, UK,
  August 29 - September 2, 2011. Proceedings}, pages 1--11, 2011.

\bibitem{Fanello14}
Sean~Ryan Fanello, Cem Keskin, Shahram Izadi, Pushmeet Kohli, David Kim, David
  Sweeney, Antonio Criminisi, Jamie Shotton, Sing~Bing Kang, and Tim Paek.
\newblock Learning to be a depth camera for close-range human capture and
  interaction.
\newblock {\em ACM Trans. Graph.}, 33(4):86:1--86:11, July 2014.

\bibitem{Qian14}
Chen Qian, Xiao Sun, Yichen Wei, Xiaoou Tang, and Jian Sun.
\newblock Realtime and robust hand tracking from depth.
\newblock In {\em 2014 {IEEE} Conference on Computer Vision and Pattern
  Recognition, {CVPR} 2014, Columbus, OH, USA, June 23-28, 2014}, pages
  1106--1113, 2014.

\bibitem{Tang14}
Danhang Tang, Hyung~Jin Chang, A.~Tejani, and Tae-Kyun Kim.
\newblock Latent regression forest: Structured estimation of 3d articulated
  hand posture.
\newblock In {\em Computer Vision and Pattern Recognition (CVPR), 2014 IEEE
  Conference on}, pages 3786--3793, June 2014.

\bibitem{Sun15}
Xiao Sun, Yichen Wei, Shuang Liang, Xiaoou Tang, and Jian Sun.
\newblock Cascaded hand pose regression.
\newblock June 2015.

\bibitem{Oberweger15}
Markus Oberweger, Paul Wohlhart, and Vincent Lepetit.
\newblock Hands deep in deep learning for hand pose estimation.
\newblock {\em CoRR}, abs/1502.06807, 2015.

\bibitem{Erol07}
Ali Erol, George Bebis, Mircea Nicolescu, Richard~D. Boyle, and Xander Twombly.
\newblock Vision-based hand pose estimation: A review.
\newblock {\em Comput. Vis. Image Underst.}, 108(1-2):52--73, October 2007.

\bibitem{SupancicRYSR15}
James Steven~Supancic III, Gr{\'{e}}gory Rogez, Yi~Yang, Jamie Shotton, and
  Deva Ramanan.
\newblock Depth-based hand pose estimation: methods, data, and challenges.
\newblock {\em CoRR}, abs/1504.06378, 2015.

\bibitem{Pavlovic97}
Vladimir~I. Pavlovic, Rajeev Sharma, and Thomas~S. Huang.
\newblock Visual interpretation of hand gestures for human-computer
  interaction: A review.
\newblock {\em IEEE Trans. Pattern Anal. Mach. Intell.}, 19(7):677--695, July
  1997.

\bibitem{Wu01}
Ying Wu and T.S. Huang.
\newblock Hand modeling, analysis and recognition.
\newblock {\em Signal Processing Magazine, IEEE}, 18(3):51--60, May 2001.

\bibitem{Athitsos03}
Vassilis Athitsos and Stan Sclaroff.
\newblock Estimating 3d hand pose from a cluttered image.
\newblock In {\em 2003 {IEEE} Computer Society Conference on Computer Vision
  and Pattern Recognition {(CVPR} 2003), 16-22 June 2003, Madison, WI, {USA}},
  pages 432--442, 2003.

\bibitem{Yang02}
Ming-Hsuan Yang, D.~Kriegman, and N.~Ahuja.
\newblock Detecting faces in images: a survey.
\newblock {\em Pattern Analysis and Machine Intelligence, IEEE Transactions
  on}, 24(1):34--58, Jan 2002.

\bibitem{Pajdla04}
AntonisA. Argyros and ManolisI.A. Lourakis.
\newblock Real-time tracking of multiple skin-colored objects with a possibly
  moving camera.
\newblock In Tomáš Pajdla and Jiří Matas, editors, {\em Computer Vision -
  ECCV 2004}, volume 3023 of {\em Lecture Notes in Computer Science}, pages
  368--379. Springer Berlin Heidelberg, 2004.

\bibitem{Chai98}
D.~Chai and K.N. Ngan.
\newblock Locating facial region of a head-and-shoulders color image.
\newblock In {\em Automatic Face and Gesture Recognition, 1998. Proceedings.
  Third IEEE International Conference on}, pages 124--129, Apr 1998.

\bibitem{Jones99}
M.J. Jones and J.M. Rehg.
\newblock Statistical color models with application to skin detection.
\newblock In {\em Computer Vision and Pattern Recognition, 1999. IEEE Computer
  Society Conference on.}, volume~1, pages --280 Vol. 1, 1999.

\bibitem{Saxe96}
D.~Saxe and R.~Foulds.
\newblock Toward robust skin identification in video images.
\newblock In {\em Automatic Face and Gesture Recognition, 1996., Proceedings of
  the Second International Conference on}, pages 379--384, Oct 1996.

\bibitem{Terrillon00}
J.-C. Terrillon, M.N. Shirazi, H.~Fukamachi, and S.~Akamatsu.
\newblock Comparative performance of different skin chrominance models and
  chrominance spaces for the automatic detection of human faces in color
  images.
\newblock In {\em Automatic Face and Gesture Recognition, 2000. Proceedings.
  Fourth IEEE International Conference on}, pages 54--61, 2000.

\bibitem{Shotton11}
J.~Shotton, A.~Fitzgibbon, M.~Cook, T.~Sharp, M.~Finocchio, R.~Moore,
  A.~Kipman, and A.~Blake.
\newblock Real-time human pose recognition in parts from single depth images.
\newblock In {\em Computer Vision and Pattern Recognition (CVPR), 2011 IEEE
  Conference on}, pages 1297--1304, June 2011.

\bibitem{Keskin12}
Cem Keskin, Furkan K\$\#305;ra\$\#231;, Yunus~Emre Kara, and Lale Akarun.
\newblock Hand pose estimation and hand shape classification using
  multi-layered randomized decision forests.
\newblock In {\em Proceedings of the 12th European Conference on Computer
  Vision - Volume Part VI}, ECCV'12, pages 852--863, Berlin, Heidelberg, 2012.
  Springer-Verlag.

\bibitem{Baak13}
Andreas Baak, Meinard Müller, Gaurav Bharaj, Hans-Peter Seidel, and Christian
  Theobalt.
\newblock A data-driven approach for real-time full body pose reconstruction
  from a depth camera.
\newblock In {\em Consumer Depth Cameras for Computer Vision}, Advances in
  Computer Vision and Pattern Recognition, pages 71--98. Springer London, 2013.

\bibitem{Plagemann10}
C.~Plagemann, V.~Ganapathi, D.~Koller, and S.~Thrun.
\newblock Real-time identification and localization of body parts from depth
  images.
\newblock In {\em Robotics and Automation (ICRA), 2010 IEEE International
  Conference on}, pages 3108--3113, May 2010.

\bibitem{Unzueta08}
Luis Unzueta, Manuel Peinado, Ronan Boulic, and \'{A}ngel Suescun.
\newblock Full-body performance animation with sequential inverse kinematics.
\newblock {\em Graph. Models}, 70(5):87--104, September 2008.

\bibitem{Sun12}
Min Sun, P.~Kohli, and J.~Shotton.
\newblock Conditional regression forests for human pose estimation.
\newblock In {\em Computer Vision and Pattern Recognition (CVPR), 2012 IEEE
  Conference on}, pages 3394--3401, June 2012.

\bibitem{ShottonJungle13}
Jamie Shotton, Toby Sharp, Pushmeet Kohli, Sebastian Nowozin, John~M. Winn, and
  Antonio Criminisi.
\newblock Decision jungles: Compact and rich models for classification.
\newblock In {\em Advances in Neural Information Processing Systems 26: 27th
  Annual Conference on Neural Information Processing Systems 2013. Proceedings
  of a meeting held December 5-8, 2013, Lake Tahoe, Nevada, United States.},
  pages 234--242, 2013.

\bibitem{Canny86}
J~Canny.
\newblock A computational approach to edge detection.
\newblock {\em IEEE Trans. Pattern Anal. Mach. Intell.}, 8(6):679--698, June
  1986.

\bibitem{Taylor14}
J.~Taylor, R.~Stebbing, V.~Ramakrishna, C.~Keskin, J.~Shotton, S.~Izadi,
  A.~Hertzmann, and A.~Fitzgibbon.
\newblock User-specific hand modeling from monocular depth sequences.
\newblock In {\em Computer Vision and Pattern Recognition (CVPR), 2014 IEEE
  Conference on}, pages 644--651, June 2014.

\bibitem{Rehg94}
J.M. Rehg and T.~Kanade.
\newblock Digiteyes: vision-based hand tracking for human-computer interaction.
\newblock In {\em Motion of Non-Rigid and Articulated Objects, 1994.,
  Proceedings of the 1994 IEEE Workshop on}, pages 16--22, Nov 1994.

\bibitem{ouhaddi99}
Hocine Ouhaddi and Patrick Horain.
\newblock 3d hand gesture tracking by model registration.
\newblock {\em Workshop on Synthetic-Natural Hybrid Coding and Three
  Dimensional Imaging}, pages 70--73, 1999.

\bibitem{Nirei96}
K.~Nirei, H.~Saito, M.~Mochimaru, and S.~Ozawa.
\newblock Human hand tracking from binocular image sequences.
\newblock In {\em Industrial Electronics, Control, and Instrumentation, 1996.,
  Proceedings of the 1996 IEEE IECON 22nd International Conference on},
  volume~1, pages 297--302 vol.1, Aug 1996.

\bibitem{Bray04}
M.~Bray, E.~Koller-Meier, P.~Muller, L.~Van~Gool, and N.N. Schraudolph.
\newblock 3d hand tracking by rapid stochastic gradient descent using a
  skinning model.
\newblock In {\em Visual Media Production, 2004. (CVMP). 1st European
  Conference on}, pages 59--68, March 2004.

\bibitem{Stenger01}
B.~Stenger, P.R.S. Mendonca, and R.~Cipolla.
\newblock Model-based 3d tracking of an articulated hand.
\newblock In {\em Computer Vision and Pattern Recognition, 2001. CVPR 2001.
  Proceedings of the 2001 IEEE Computer Society Conference on}, volume~2, pages
  II--310--II--315 vol.2, 2001.

\bibitem{Kennedy95}
James Kennedy and Russell~C. Eberhart.
\newblock Particle swarm optimization.
\newblock In {\em Proceedings of the IEEE International Conference on Neural
  Networks}, pages 1942--1948, 1995.

\bibitem{Kennedy01}
James Kennedy, Russell~C Eberhart, and Y~Shi.
\newblock Swarm intelligence. 2001.
\newblock {\em Kaufmann, San Francisco}, 1:700--720, 2001.

\bibitem{Besl92}
Paul~J. Besl and Neil~D. McKay.
\newblock A method for registration of 3-d shapes.
\newblock {\em IEEE Trans. Pattern Anal. Mach. Intell.}, 14(2):239--256,
  February 1992.

\bibitem{Rusinkiewicz01}
Szymon Rusinkiewicz and Marc Levoy.
\newblock Efficient variants of the {ICP} algorithm.
\newblock In {\em Third International Conference on 3D Digital Imaging and
  Modeling (3DIM)}, June 2001.

\bibitem{Pellegrini08}
S.~Pellegrini, K.~Schindler, and D.~Nardi.
\newblock A generalisation of the icp algorithm for articulated bodies.
\newblock In {\em Proceedings of the British Machine Vision Conference}, pages
  87.1--87.10. BMVA Press, 2008.
\newblock doi:10.5244/C.22.87.

\bibitem{Goldberg89}
David~E. Goldberg.
\newblock {\em Genetic Algorithms in Search, Optimization and Machine
  Learning}.
\newblock Addison-Wesley Longman Publishing Co., Inc., Boston, MA, USA, 1st
  edition, 1989.

\bibitem{Juang04}
Chia-Feng Juang.
\newblock A hybrid of genetic algorithm and particle swarm optimization for
  recurrent network design.
\newblock {\em Trans. Sys. Man Cyber. Part B}, 34(2):997--1006, April 2004.

\bibitem{Yasuda10}
T.~Yasuda, K.~Ohkura, and Y.~Matsumura.
\newblock Extended pso with partial randomization for large scale multimodal
  problems.
\newblock In {\em World Automation Congress (WAC), 2010}, pages 1--6, Sept
  2010.

\bibitem{Tseng99}
Paul Tseng.
\newblock Fortified-descent simplicial search method: A general approach.
\newblock {\em SIAM Journal on Optimization}, 10(1):269--288, 1999.

\bibitem{WuLin01}
Ying Wu, J.Y. Lin, and T.S. Huang.
\newblock Capturing natural hand articulation.
\newblock In {\em Computer Vision, 2001. ICCV 2001. Proceedings. Eighth IEEE
  International Conference on}, volume~2, pages 426--432 vol.2, 2001.

\bibitem{RogezSKMR14}
Gr{\'{e}}gory Rogez, James Steven~Supancic III, Maryam Khademi, Jos{\'{e}}
  Mar{\'{\i}}a~Mart{\'{\i}}nez Montiel, and Deva Ramanan.
\newblock 3d hand pose detection in egocentric {RGB-D} images.
\newblock {\em CoRR}, abs/1412.0065, 2014.

\bibitem{libhand}
Marin \v{S}ari\'{c}.
\newblock Libhand: A library for hand articulation, 2011.
\newblock Version 0.9.

\bibitem{Melax13}
Stan Melax, Leonid Keselman, and Sterling Orsten.
\newblock Dynamics based 3d skeletal hand tracking.
\newblock In {\em Proceedings of Graphics Interface 2013}, GI '13, pages
  63--70, Toronto, Ont., Canada, Canada, 2013. Canadian Information Processing
  Society.

\bibitem{Everingham10}
Mark Everingham, Luc Van~Gool, ChristopherK.I. Williams, John Winn, and Andrew
  Zisserman.
\newblock The pascal visual object classes (voc) challenge.
\newblock {\em International Journal of Computer Vision}, 88(2):303--338, 2010.

\bibitem{Russakovsky15}
Olga Russakovsky, Jia Deng, Hao Su, Jonathan Krause, Sanjeev Satheesh, Sean Ma,
  Zhiheng Huang, Andrej Karpathy, Aditya Khosla, Michael Bernstein,
  Alexander~C. Berg, and Li~Fei-Fei.
\newblock {ImageNet Large Scale Visual Recognition Challenge}.
\newblock {\em International Journal of Computer Vision (IJCV)}, 2015.

\end{thebibliography}
\end{document}